%% file: main.tex
\documentclass[10pt,twocolumn,letterpaper]{article}

\usepackage[pagenumbers]{cvpr} 
\usepackage{multirow}
\usepackage{graphicx}
\usepackage{booktabs}
\usepackage{colortbl}
\usepackage[accsupp]{axessibility}
\usepackage{url}
\usepackage{pifont}
\usepackage{marvosym}

\definecolor{cvprblue}{rgb}{0.21,0.49,0.74}
\usepackage[pagebackref,breaklinks,colorlinks,allcolors=cvprblue]{hyperref}

\def\paperID{6289} 
\def\confName{CVPR}
\def\confYear{2025}

\title{HRAvatar: High-Quality and Relightable Gaussian Head Avatar}
\makeatletter
\newcommand{\printfnsymbol}[1]{%
  \textsuperscript{\@fnsymbol{#1}}%
}
\renewcommand*{\@fnsymbol}[1]{\ensuremath{\ifcase#1\or *\or \dagger\or \ddagger\or
   \mathsection\or \mathparagraph\or \|\or **\or \dagger\dagger
   \or \ddagger\ddagger \else\@ctrerr\fi}}

\author{
    Dongbin Zhang$^{1,2}$\thanks{Intern at IDEA} \quad Yunfei Liu$^{2}$ \quad Lijian Lin$^{2}$ \quad Ye Zhu$^{2}$ \quad Kangjie Chen$^{1}$ 
   \\ Minghan Qin$^{1}$ \quad Yu Li$^{2}$\thanks{Corresponding authors} \quad Haoqian Wang$^{1}$\printfnsymbol{2} \\
    $^1$Tsinghua Shenzhen International Graduate School, Tsinghua University \\
    $^2$International Digital Economy Academy (IDEA) 
}

\begin{document}
\twocolumn[{%
\renewcommand\twocolumn[1][]{#1}%
\maketitle
\begin{center}
    \centering
    \captionsetup{type=figure}
    \includegraphics[width=1\textwidth]{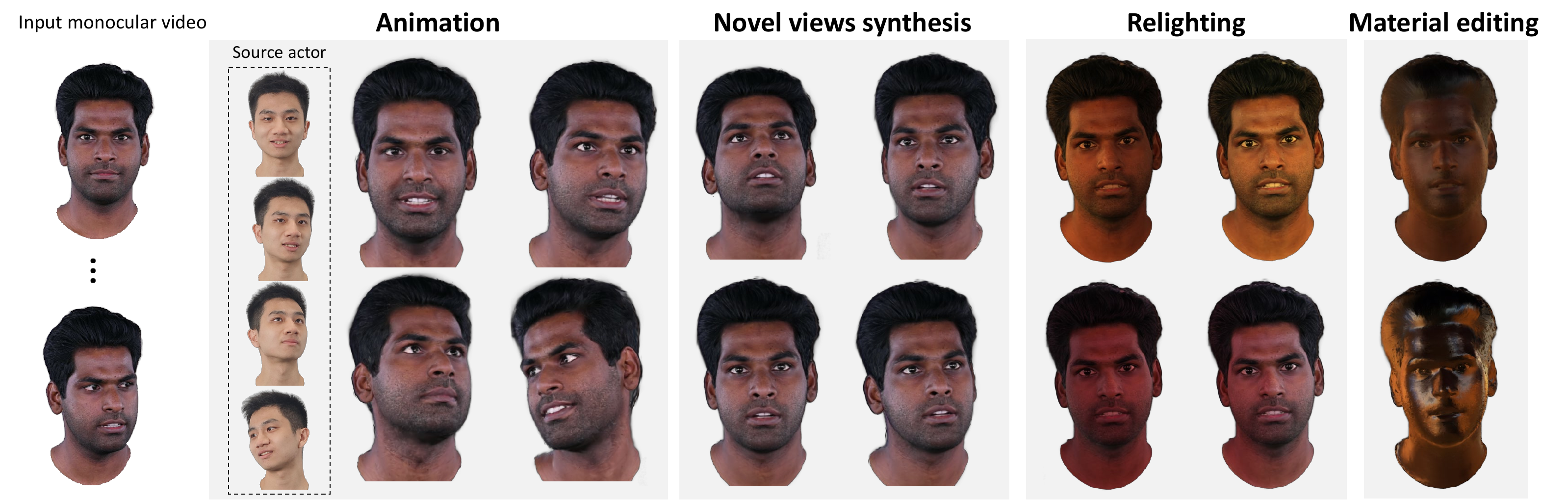}
    \captionof{figure}{With monocular video input, HRAvatar reconstructs a high-quality, animatable 3D head avatar that enables realistic relighting effects and simple material editing.
    }
    \label{teaser}
\vspace{-6pt}
\end{center}%
}]

{\let\thefootnote\relax\footnotetext{{$^{*}$ Intern at IDEA. ~ $^{\dagger}$ Corresponding authors.}}}

\maketitle
\input{sec/0_abstract}

\input{sec/1_intro}
\input{sec/2_relatedwork}

\input{sec/3_method}

\input{sec/4_expriment}

\input{sec/5_discussion}

{
    \small
    \bibliographystyle{ieeenat_fullname}
    \bibliography{main}
}

\input{sec/X_suppl}

\end{document}

%% file: sec/0_abstract.tex
\begin{abstract}
Reconstructing animatable and high-quality 3D head avatars from monocular videos, especially with realistic relighting, is a valuable task. However, the limited information from single-view input, combined with the complex head poses and facial movements, makes this challenging. Previous methods achieve real-time performance by combining 3D Gaussian Splatting with a parametric head model, but the resulting head quality suffers from inaccurate face tracking and limited expressiveness of the deformation model. These methods also fail to produce realistic effects under novel lighting conditions. To address these issues, we propose HRAvatar, a 3DGS-based method that reconstructs high-fidelity, relightable 3D head avatars. HRAvatar reduces tracking errors through end-to-end optimization and better captures individual facial deformations using learnable blendshapes and learnable linear blend skinning. Additionally, it decomposes head appearance into several physical properties and incorporates physically-based shading to account for environmental lighting. Extensive experiments demonstrate that HRAvatar not only reconstructs superior-quality heads but also achieves realistic visual effects under varying lighting conditions. Video results and code are available at the \href{https://eastbeanzhang.github.io/HRAvatar/}{\textcolor{magenta}{project page}}.
\end{abstract}

%% file: sec/1_intro.tex
\section{Introduction}
\label{intro}
Creating a 3D head avatar is essential for film, gaming, immersive meetings, AR/VR, etc.
In these applications, the avatar must meet several requirements: animatable, real-time, high-quality, and visually realistic. However, achieving a highly realistic and animatable head avatar from widely-used monocular video remains challenging. 

Research in this area spans many years. Early efforts \citep{li2017learning,paysan20093d,cao2013facewarehouse} develop parametric head models based on 3D Morphable Models (3DMM) theory~\citep{10.1145/311535.311556}. These methods allow registering 3D head scans to parametric models for 3D facial mesh reconstruction. With the rise of deep learning, methods \citep{liu2025teaser,chang2017faceposenet,danvevcek2022emoca,zielonka2022towards} use parametric model priors to simplify head mesh reconstruction from videos, either through estimation or frame-wise optimization, \textit{i.e.}, 3D face tracking. While these methods generalize well for expressions and pose variations, their fixed topology limits complex hair modeling and fine-grained appearance reconstruction. To address this issue, some researchers have turned to Neural Radiance Fields (NeRF) \citep{mildenhall2020nerf} for modeling head avatars \citep{xu2023avatarmav,xu2023latentavatar,zhao2023havatar,qin2024high}. These approaches enable complete geometry and appearance reconstruction, including hair, glasses, earrings, \textit{etc.} However, they are limited by slow rendering and long training time. Recently, 3D Gaussian Splatting (3DGS) \citep{kerbl20233d} has gained significant attention for its fast rendering speed. Some methods \citep{xiang2024flashavatar,shao2024splattingavatar,chen2024monogaussianavatar} have extended 3DGS to head avatar reconstruction, significantly improving rendering speed compared to NeRF-based methods.

Although previous 3DGS-based methods have made progress in animatability and real-time rendering, their reconstruction quality is constrained by two major factors: \textbf{limited deformation flexibility} and \textbf{inaccurate expression tracking}. Additionally, they are \textbf{unable to produce realistic relighting effects}. Specifically, our motivation primarily stems from the following three points. \textbf{1)} Head reconstruction requires a geometric model to deform from the compact canonical space to various states based on different expressions and poses. Recent methods~\citep{xiang2024flashavatar, shao2024splattingavatar} model geometric deformations of Gaussian points by rigging them to universal parametric model mesh faces. However, parametric models may not accurately capture personalized deformations. \textbf{2)} Before training, these methods extract FALME parameters by fitting pseudo-2D facial keypoints, which are usually error-prone and lead to suboptimal results. Methods like PointAvatar \citep{zheng2023pointavatar} try to directly optimize these parameters during training. Such a design may introduce a mismatch from pre-tracked parameters and limit generalization to new expressions and poses. Consequently, such methods still require post-optimization during testing. \textbf{3)} Under monocular and unknown lighting settings, existing 3DGS-based methods directly fit the colors of the avatar, causing an inability to relight and mix the person's intrinsic appearance with ambient lighting.

To tackle the aforementioned challenges, we propose HRAvatar, which utilizes 3D Gaussian points for high-quality head avatar reconstruction with realistic relighting from monocular videos, as \cref{teaser}. We propose a learnable blendshapes and learnable linear blend skinning strategy, allowing the Gaussian points for flexible deformation from canonical space to pose space. Additionally, we utilize an expression encoder to extract accurate facial expression parameters in an end-to-end training manner, which not only reduces the impact of tracking errors on reconstruction but also ensures the generalization of expression parameters estimation. To achieve realistic and real-time relighting, we model the head's appearance by using albedo, roughness, Fresnel reflectance, \etc with an approximate physically-based shading model. An albedo pseudo-prior is also employed to better decouple the albedo. For a detailed comparison and distinction from previous methods, please refer to the supporting materials. Benefiting from these techniques, HRAvatar can reconstruct fine-grained and expressive avatars while achieving realistic relighting effects. 

In summary:
\textbf{a)} We present HRAvatar, a method for monocular reconstruction of head avatars using 3D Gaussian points. HRAvatar leverages learnable blendshapes and learnable linear blend skinning for flexible and precise geometric deformations, with a precise expression encoder reducing tracking errors for high-quality reconstructions.
\textbf{b)} We incorporate intrinsic priors to model head appearance under unknown lighting conditions. Combined with a physically-based shading model, we achieve realistic lighting effects across different environments.
\textbf{c)} Experimental results demonstrate that HRAvatar outperforms existing methods in overall quality, enabling realistic relighting in real-time and simple material editing.

%% file: sec/2_relatedwork.tex
\begin{figure*}[tb]
  \centering
  \includegraphics[width=\textwidth]{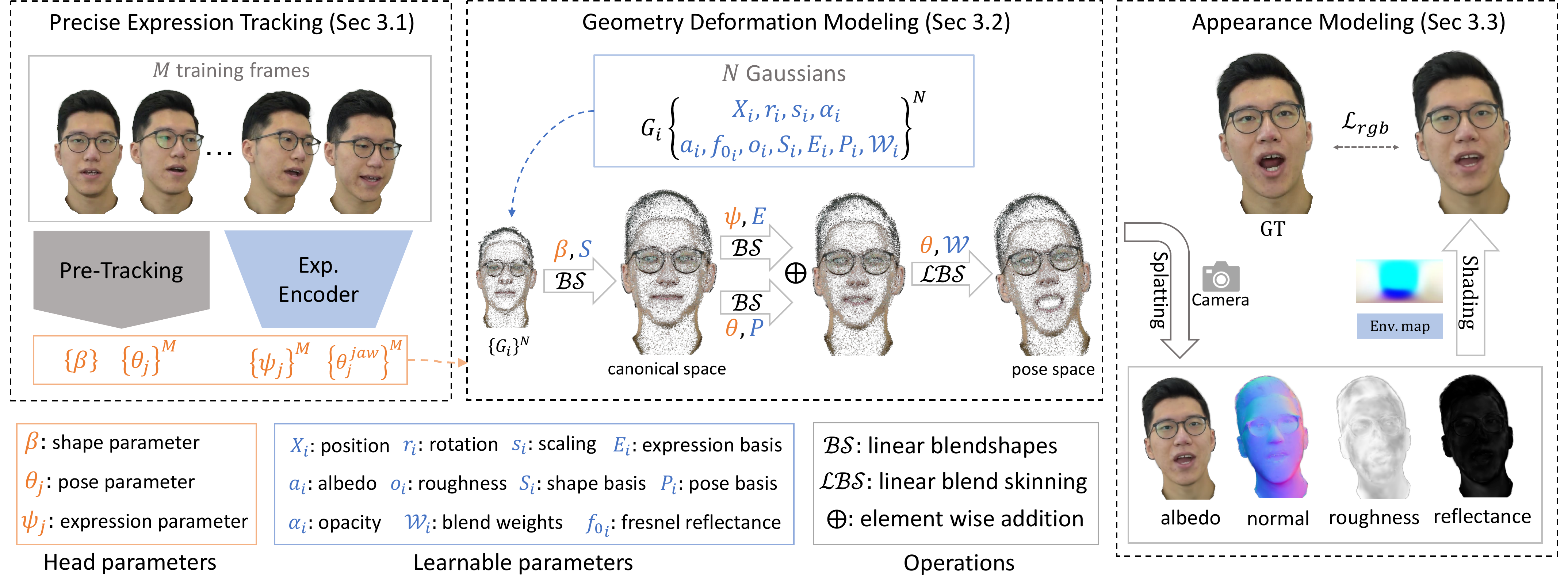}

  \caption{Given a monocular video with unknown lighting and $M$ frames, we first track fixed shape parameter $\beta$ and pose parameters $\{\theta_j\}^M$ through iterative optimization before training. Expression parameters $\{\psi_j\}^M$ and jaw poses $\theta^{jaw}$ are estimated via an expression encoder, which is optimized during training. With these parameters, we transform the Gaussian points into pose space using learnable linear blendshapes $\mathcal{BS}$ and linear blend skinning $\mathcal{LBS}$. We then render the Gaussian points to obtain albedo, roughness, reflectance, and normal maps. Finally, we compute pixel colors using physically-based shading with optimizable environment maps.
  }
  \label{pipeline}
\end{figure*}

\section{Related Work}
\label{Relatedwork}

\subsection{3D Radiance Fields}
Image-based 3D reconstruction has become a vibrant research area due to its photorealistic visuals. NeRF \citep{mildenhall2020nerf} introduced a novel method using MLPs to represent a 3D scene as a continuous density and color field, enabling differentiable image rendering through volume rendering. This approach has inspired numerous follow-up studies \citep{martin2021nerf,yu2021pixelnerf,barron2021mip,wang2021neus,chu2024gpavatar}. However, NeRF faces heavy computational challenges due to extensive MLP queries. Instant-NGP \citep{muller2022instant} employs multi-resolution hash encoding to accelerate inference. Additionally, some methods, propose hybrid 3D representations \citep{chan2022efficient,cao2023hexplane,fridovich2023k} to improve efficiency.
Recently, 3DGS introduces an explicit representation using Gaussian points, achieving real-time rendering with an efficient tile-based rasterizer. It rapidly gains attention, and researchers applying it to various fields \citep{wu20244d,qin2024langsplat,zhang2024gaussian,yu2024mip,charatan2024pixelsplat,huang20242d,kirschstein2024gghead,chen2024slgaussian} to exploit its efficiency. Our work also builds upon 3DGS to achieve real-time rendering.

\subsection{3D Head Reconstruction}
\noindent\textbf{Geometric mesh reconstruction.} Traditional 3DMM \citep{10.1145/311535.311556} uses Principal Component Analysis (PCA) to create a parameterized facial model that represents appearance and geometric variations in a linear space. BFM \citep{paysan20093d} improves on this by adding more scanned facial data, resulting in a richer model. FLAME \citep{li2017learning} introduces extra joints for the eyes, jaw, and neck, enabling more realistic facial motion. DECA \citep{feng2021learning} builds on FLAME by estimating parameters like shape and pose from a single image and capturing finer wrinkles. SMIRK \citep{retsinas20243d} enhances tracking accuracy by using an image-to-image module to provide more precise supervision signals. Besides geometry, some works \citep{lattas2020avatarme,towards_monorefl,caselles2023sira, MoSAR} also focus on learning intrinsic attributes for relightable mesh reconstruction from a single image.
 
\noindent\textbf{Image-based head reconstruction.} Recent advances in neural radiance fields combine 3DMM for view-consistent, photorealistic 3D head reconstruction, which can be generally divided into two categories. \textit{Multi-view-based methods}. Some studies explore multi-view video-based head \citep{lombardi2021mixture,xu2024gaussian,qian2024gaussianavatars,giebenhain2024npga,teotia2024gaussianheads} and full-body \cite{li2024animatablegaussians,li2023posevocab,liao2024vinecs} reconstruction. However, these approaches require multiple synchronized cameras, making them more complex and less convenient than single-phone captures. Although multi-view-based methods can achieve impressive results, their setup limits the applicability of these approaches. \textit{Monocular-based methods}.  NeRFace \citep{Gafni_2021_CVPR} extends NeRF to dynamic forms by incorporating expression and pose parameters as conditional inputs, enabling animatable head reconstruction. IMavatar \citep{zheng2022avatar} models deformation fields for expression and pose motions, using iterative root-finding to locate the canonical surface intersection for each pixel. Point-avatar \citep{zheng2023pointavatar} introduces a novel point-based representation for more efficient animatable avatars. While Point-avatar learns person-specific deformation fields through a shared MLP, our method independently learns per-point blendshapes basis and blend weights, leading to a more flexible deformation modeling. INSTA \citep{zielonka2023instant} speeds up training by using multi-resolution hashing for 3D head representation. Recent works \citep{shao2024splattingavatar,xiang2024flashavatar} based on 3DGS achieve significant breakthroughs in rendering speed. 3D Gaussian Blendshapes (GBS) \citep{ma20243d} learn Gaussian basis to better handle expression movements but struggle with pose variations. In contrast, our method utilizes learnable linear blend skinning for flexible point pose transformations, enabling better handling of person-specific head pose animation, while also providing realistic relighting effects.

\subsection{Neural Relighting}
Implementing relighting in reconstructed 3D scenes is difficult. For static scenes, some methods \citep{zhang2021neural,gao2020deferred,xu2023renerf} use learning-based approaches to learn relightable appearances from images under varying lighting. In contrast, inverse rendering methods \citep{zhang2021nerfactor,zhang2021physg,boss2021nerd,zhang2022invrender} leverage reflection models like BRDF for more realistic relighting. Recent works \citep{gao2023relightable,jiang2024gaussianshader} integrate BRDF into 3DGS and methods \citet{wu2024deferredgs,ye20243d} introduce deferred shading for efficient relighting or specular rendering of static scenes. While simplified physical rendering models can be inaccurate, many methods \citep{wu2024deferredgs,jin2023tensoir,li2024animatable} add fitting-based rendering branches to improve reconstruction results. Although some researchers combine physical reflection models with dynamic radiance fields to achieve relightable head avatars \citep{li2022eyenerf,yang2024vrmm,saito2024relightable}, they require data under controlled lighting conditions. Reconstructing relightable 3D head avatars under monocular unknown lighting is still underexplored. Point-avatar models lighting but relies on trained shading networks, unable to flexibly relight through environment maps. Unlike NeRF or 3DGS, FLARE \cite{bharadwaj2023flare} reconstructs avatars with meshes and uses a BRDF for relighting, but the reconstruction quality is limited. Our method not only reconstructs superior head avatars but also supports realistic and real-time relighting.

%% file: sec/3_method.tex
\section{Method}
\label{Method}
As mentioned, previous methods for head reconstruction suffer from inaccurate 3D expression tracking and limited person-specific deformation. They also cannot achieve realistic relighting effects. To tackle these challenges, we enhance expression tracking through end-to-end optimization (\cref{face tracking}). We also adopt learning strategy for both linear blendshapes and blend skinning for more flexible deformation of Gaussian points (\cref{deformation modeling}). Physically-based shading is employed to realistically model head appearance, which makes our model achieve realistic relighting (\cref{Appearance modeling}). The overall pipeline is illustrated in \cref{pipeline}.

\subsection{Precise Expression Tracking}
\label{face tracking}
Although existing face tracking methods can accurately track head pose and shape parameters, they often struggle to precisely estimate expression parameters. Since these parameters control head expressions, inaccuracies can cause deformation errors, compromising reconstruction quality.
To mitigate this issue while maintaining good generalization, we propose to use an expression encoder $\mathcal{E}$ to extract more accurate expression parameters, which is end-to-end trained with subsequent 3D avatar reconstruction:
\begin{equation}
    {\psi,\theta^{jaw}}=\mathcal{E}(I),
    \label{eq4}
\end{equation}
where $\psi$ and $\theta^{jaw}$ represent the expression and jaw pose parameters, respectively. Note that traditional fitting-based methods optimize face parameters using pseudo labels (e.g., pre-estimated 2D landmarks). In contrast, our encoder is trained end-to-end during reconstruction, utilizing photometric loss with ground-truth face images for supervision. Hence, the proposed encoder enables more precise expression tracking and maintains good generalization.

Since point transformations are sensitive to jaw pose parameters \cite{li2017learning}, we introduce a regularization loss that constrains the distance between the inferred and pre-tracked jaw poses $\hat{\theta}^{jaw}$:
\begin{equation}
\mathcal{L}_{jaw}=\left\|\hat{\theta}^{jaw}-\theta^{jaw}\right\|_2.
    \label{eq5}
\end{equation}
Other pose parameters in $\theta$ and shape parameters $\beta$ are pre-tracked using \cite{zheng2022avatar}, with 
$\beta$ shared across all frames.

\subsection{Geometry Deformation Modeling}
\label{deformation modeling}
Like most methods, we employ a deformation model to map points from canonical space to pose space based on expression and pose parameters. However, facial shapes, expressions, and pose deformations vary widely among individuals, making it difficult for parametric head models to accurately recover each person's unique shape and deformations. To address this, we independently learn per-point blendshapes basis and blend weights adaptively for more flexible geometric deformation.

\noindent\textbf{Learnable linear blendshapes.}
Similar to FLAME \citep{li2017learning}, we use linear blendshapes to model geometric displacement. For each Gaussian point, we introduce three additional attributes: shape basis ${S}=\{{S}^1,...,{S}^{|\beta|}\}\in \mathbb{R}^{N\times3\times|\beta|}$, expression basis ${E}=\{{E}^1,...,{E}^{|\psi|}\}\in \mathbb{R}^{N\times3\times|\psi|}$ and pose basis ${P}=\{{P}^1,...,{P}^{9K}\}\in \mathbb{R}^{N\times3\times9K}$. These are learnable parameters that fit the individual head shape and deformations. First, we compute the shape offset to displace the points to the canonical space $X_{c}$ using shape blendshapes:
\begin{equation}
     \mathcal{BS}({\beta},{S})=\sum_{m=1}^{|\beta|}\beta^m{S}^m,~X_{c}=X+\mathcal{BS}(\beta,{S}),
    \label{eq6}
\end{equation}
where $\mathcal{BS}$ denotes linear blendshapes and $\beta=\{\beta^1,...,\beta^{|\beta|}\}\in\mathbb{R}^{|\beta|}$ is the shape parameter. Next, we compute expression and pose offsets in the same manner, using expression blendshapes and pose blendshapes to model facial expressions:
\begin{equation}
     X_{e}=X_c+\mathcal{BS}(\psi,{E})+\mathcal{BS}(\mathcal{R}(\theta^*)-\mathcal{R}(\theta^0),{P}) ,
    \label{eq7}
\end{equation}
where $\psi=\{\psi^1,...,\psi^{|\psi|}\}\in\mathbb{R}^{|\psi|}$ is the expression parameter, and $\theta\in\mathbb{R}^{3(K+1)}$ is the pose parameter representing the axis-angle rotation of the points relative to the joints. $\theta^*$ excludes the global joint, with $K=4$. $\mathcal{R}(\theta)$ is the flattened rotation matrix vector obtained by Rodrigues' formula, and $\theta^o$ represents zero pose.

\noindent\textbf{Learnable linear blend skinning.} After applying linear displacement, we transform Gaussian points into pose space using Linear Blend Skinning (LBS). Each Gaussian point is assigned with a learnable blend weight attribute $\mathcal{W}\in\mathbb{R}^{N\times K}$ to accommodate individual pose deformations. $\mathcal{LBS}$ rotates the points $X_e$ around each joints $\mathcal{J}(\beta)$ and linearly weighted by $\mathcal{W}$, defined as:
\begin{equation}
     X_p=\mathcal{LBS}(X_e,\mathcal{J}(\beta),\mathcal{W})={R}_{lbs}X_e+T_{lbs},
    \label{eq8}
\end{equation}
where $\mathcal{J}(\beta)\in\mathbb{R}^{K\times3}$ represents the positions of the neck, jaw, and eyeball joints. To maintain geometric consistency, the rotation attributes of the Gaussians are also transformed by the weighted rotation matrix ${R}_{lbs}$: $R_p={R}_{lbs}R$.

\noindent\textbf{Geometry initialization.} To facilitate easier learning, we leverage FLAME's geometric and deformation priors. We initialize the positions of the Gaussian points through linear interpolation on the FLAME mesh faces. The same method is applied to initialize the blendshapes basis and blend weights. Other geometric attributes, like rotation and scale, are initialized similarly to 3DGS.

\subsection{Appearance Modeling}
\label{Appearance modeling}
3DGS uses spherical harmonics to model the view-dependent appearance of each point, but it cannot simulate visual effects under new lighting conditions. To overcome this, we introduce a novel appearance modeling approach that decomposes the appearance into three properties: albedo $a$, roughness $o$, and Fresnel base reflectance $f_0$. We then utilize a BRDF model \citep{burley2012physically} for physically-based shading of the image. To enhance efficiency, we apply the SplitSum approximation technique \citep{karis2013real} to precompute the environment map.

\noindent\textbf{Shading.} First, we render the albedo map $\mathbf{A}$, roughness map $\mathbf{O}$, reflectance map $\mathbf{{F}_0}$, and normal map $\mathbf{N}$ using rasterizer. The specular and diffuse maps are then calculated as follows:
\begin{equation}
\begin{split}
I_{specular} &= I_{env}(\mathbf{R}, \mathbf{O}) \cdot \left( ks \cdot I_{BRDF}(\mathbf{O}, \mathbf{N} \cdot \mathbf{V})[0] \right. \\
&\quad + \left. I_{BRDF}(\mathbf{O}, \mathbf{N} \cdot \mathbf{V})[1] \right),
\label{eq9}
\end{split}
\end{equation}
\begin{equation}
I_{diffuse}= \mathbf{A} \cdot I_{irr}(\mathbf{N}),
\label{eq10}
\end{equation}
where $\mathbf{V}$ is the view direction map derived from the camera parameters and $\mathbf{R}$ is the reflection direction map, computed as $\mathbf{R}=2 (\mathbf{N}\cdot\mathbf{V})\mathbf{N}-\mathbf{V}$. $I_{BRDF}$ is a precomputed map of the simplified BRDF integral. We use an approximate Fresnel equation $\tilde{\mathcal{F}}$ to compute the specular reflectance $ks$:
\begin{equation}
\begin{split}
ks &= \tilde{\mathcal{F}}(\mathbf{N} \cdot \mathbf{V}, \mathbf{O}, \mathbf{F_0}) = \mathbf{F_0} + \left( \max\left(1 - \mathbf{O}, \mathbf{F_0}\right) \right. \\
   &\quad \left. - \mathbf{F_0} \right) \cdot 2^{\left( -5.55473 (\mathbf{N} \cdot \mathbf{V}) - 6.698316 \right) \cdot (\mathbf{N} \cdot \mathbf{V})}.
\label{eq11}
\end{split}
\end{equation}
The final shaded image is computed as: $I_{shading}=I_{diffuse}+I_{specular}$. During training, we optimize two cube maps: the environment irradiance map $I_{irr}$ and the prefiltered environment map $I_{env}$. $I_{env}(\mathbf{R},\mathbf{O})$ provides radiance values based on the reflection directions and roughness, while $I_{irr}(\mathbf{N})$ provides irradiance values based on the normal directions.

\noindent\textbf{Normal estimation.} Smooth and accurate normals are essential for physical rendering, as rough normals can cause artifacts during relighting. Following \citet{jiang2024gaussianshader}, we use the shortest axis of each Gaussian point as its normal $n$. To ensure the correct direction and geometric consistency, we supervise the rendered normal map $\mathbf{N}$ with the normal map $\hat{\mathbf{N}}$ obtained from depth derivatives:
\begin{equation}
\mathcal{L}_{normal}= \left\| \mathbf{1} - \mathbf{N}\cdot\hat{\mathbf{N}} \right\|_1 .
    \label{eq12}
\end{equation}
\textbf{Intrinsic prior}. Disentangling material properties under constant unknown lighting is challenging due to inherent uncertainties. When reconstructing heads under non-uniform lighting, local lighting effects can be erroneously coupled into the albedo, resulting in unrealistic relighting. To address this, we use an existing model \cite{chen2024intrinsicanything} to extract pseudo-ground-truth albedos $\mathbf{A}^{gt}$, supervising the rendered albedos for a more realistic appearance, as \cref{eq13}. We also constrain the roughness and base reflectance within predefined ranges: $o\in[\tau_{min}^o,\tau_{max}^o]$, $f_0\in[\tau_{min}^{f_0},\tau_{max}^{f_0}]$.
\begin{equation}
\mathcal{L}_{albedo} = \left\|\mathbf{A} - \mathbf{A}^{gt} \right\|_1  .
    \label{eq13}
\end{equation}

\subsection{Optimization}
\label{Optimization}
During optimization, we retain the point densification and pruning strategy from 3DGS, with additional attributes inherited similarly. In addition to the previously mentioned losses, we use the Mean Absolute Error ($\mathrm{MAE}$) and D-SSIM to calculate the error between the rendered image and ground truth, as \cref{eq15}. We also apply Total Variation (TV) loss $\mathcal{L}_{tv}$ to the rendered roughness map $\mathbf{O}$ to ensure smoothness. The total loss function is given in \cref{eq14}. The weights for each loss component are set as follows: $\lambda_{jaw}=0.1$,~$\lambda_1=0.8$,~$\lambda_{\mathcal{W}}=0.1$,~$\lambda_{normal}=10^{-5}$,~$\lambda_{albedo}=0.25$,~$\lambda_{tv}=0.02$.
\begin{equation}
\begin{split}
\mathcal{L}_{total}&=\mathcal{L}_{rgb}+\lambda_{jaw}\mathcal{L}_{jaw}+\lambda_{normal}\mathcal{L}_{normal}+\\
& \lambda_{albedo}\mathcal{L}_{albedo}+\lambda_{tv}\mathcal{L}_{tv}(\mathbf{O}),
    \label{eq14}
\end{split}
\end{equation}
\begin{equation}
\begin{split}
\text{where} \quad \mathcal{L}_{rgb} = \lambda_{1}\left\| I_{shading} - I_{gt} \right\|_1 +\\
(1-\lambda_{1})\mathcal{L}_\mathrm{D-SSIM}(I_{shading}, I_{gt}).
\label{eq15}
\end{split}
\end{equation}

\begin{table*}[t]
\resizebox{\linewidth}{!}{
\begin{tabular}{ccccccccccccc}
     \toprule
       \multirow{2}{*}{Method}  & \multicolumn{4}{c}{INSTA dataset} & \multicolumn{4}{c}{HDTF dataset}  &  \multicolumn{4}{c}{self-captured dataset} \\
      \cline{2-5}    \cline{6-9}     \cline{10-13} 
     & PSNR$\uparrow$ & $\text{MAE}^*$$\downarrow$ & SSIM$\uparrow$ & LPIPS$\downarrow$ & PSNR$\uparrow$ & $\text{MAE}^*$$\downarrow$ & SSIM$\uparrow$ & LPIPS$\downarrow$ & PSNR$\uparrow$ & $\text{MAE}^*$$\downarrow$ & SSIM$\uparrow$ & LPIPS$\downarrow$ \\
    \midrule
   INSTA   & 27.85 &1.309 & 0.9110 & 0.1047 & 25.03 &2.333 & 0.8475 & 0.1614 & 25.91 &1.910 & 0.8333 & 0.1833 \\
   Point-avatar & 26.84 &1.549 & 0.8970 & 0.0926 & 25.14 &2.236 & 0.8385 & 0.1278 & 25.83 &1.692 & 0.8556 & 0.1241 \\
   Splatting-avatar& 28.71 &1.200 & 0.9271 & 0.0862 & 26.66 &2.01 & 0.8611 & 0.1351 & 26.47 &1.711 & 0.8588 & 0.1550 \\
   Flash-avatar& 29.13 &1.133 & 0.9255 & 0.0719 & 27.58 &1.751 & 0.8664 & 0.1095 & 27.46 &1.632 & 0.8348 & 0.1456 \\
   GBS & 29.64 &1.020 & 0.9394 & 0.0823 & 27.81 &1.601 & 0.8915 & 0.1297 & 28.59 &1.331 & 0.8891 & 0.1560 \\
   HRAvatar (Ours)   & \bf30.36 & \bf0.845 & \bf0.9482 & \bf0.0569 & \bf28.55 & \bf1.373 & \bf0.9089 & \bf0.0825 & \bf28.97 & \bf1.123 & \bf0.9054 & \bf0.1059 \\
   \bottomrule
\end{tabular}
}
\caption{Average quantitative results on the INSTA, HDTF, and self-captured datasets. Our method outperforms others in PSNR, $\text{MAE}^*$ ($\text{MAE}\times10^2$), SSIM, and LPIPS metrics.}
\label{self-reenactment-table}
\end{table*}

\begin{table}[t]
\resizebox{\linewidth}{!}{
\begin{tabular}{ccccc}
     \toprule
       &  PSNR$\uparrow$ &  $\text{MAE}^*$$\downarrow$ & SSIM$\uparrow$ &  LPIPS$\downarrow$ \\

    \midrule
   full (ours)   & \bf 30.36 & \bf0.845 & \bf0.9482 & \underline{0.0569}  \\
  rigged to FLAME   & 29.79 & 0.937 & 0.9431 & 0.0695   \\
  MLP deform   & 29.67 & 0.966 & 0.941 & 0.0706   \\
   w/o exp. encoder  & 29.70 & 0.933 & 0.9438 & 0.0667   \\
   w/o learnable deform & 29.83 & 0.923 & 0.9440 & 0.0684  \\
   w/o PBS & \underline{30.34} & \underline{0.850} & \underline{0.9480} & \bf0.0563   \\
   \bottomrule
\end{tabular}
}
\caption{Ablation quantitative results on the INSTA dataset. \textbf{Bold} marks the best results, and \underline{underline} marks the second best results.}
\label{ablation-table}
\end{table}

\begin{figure*}[t]
  \centering
  \includegraphics[width=0.93\textwidth]{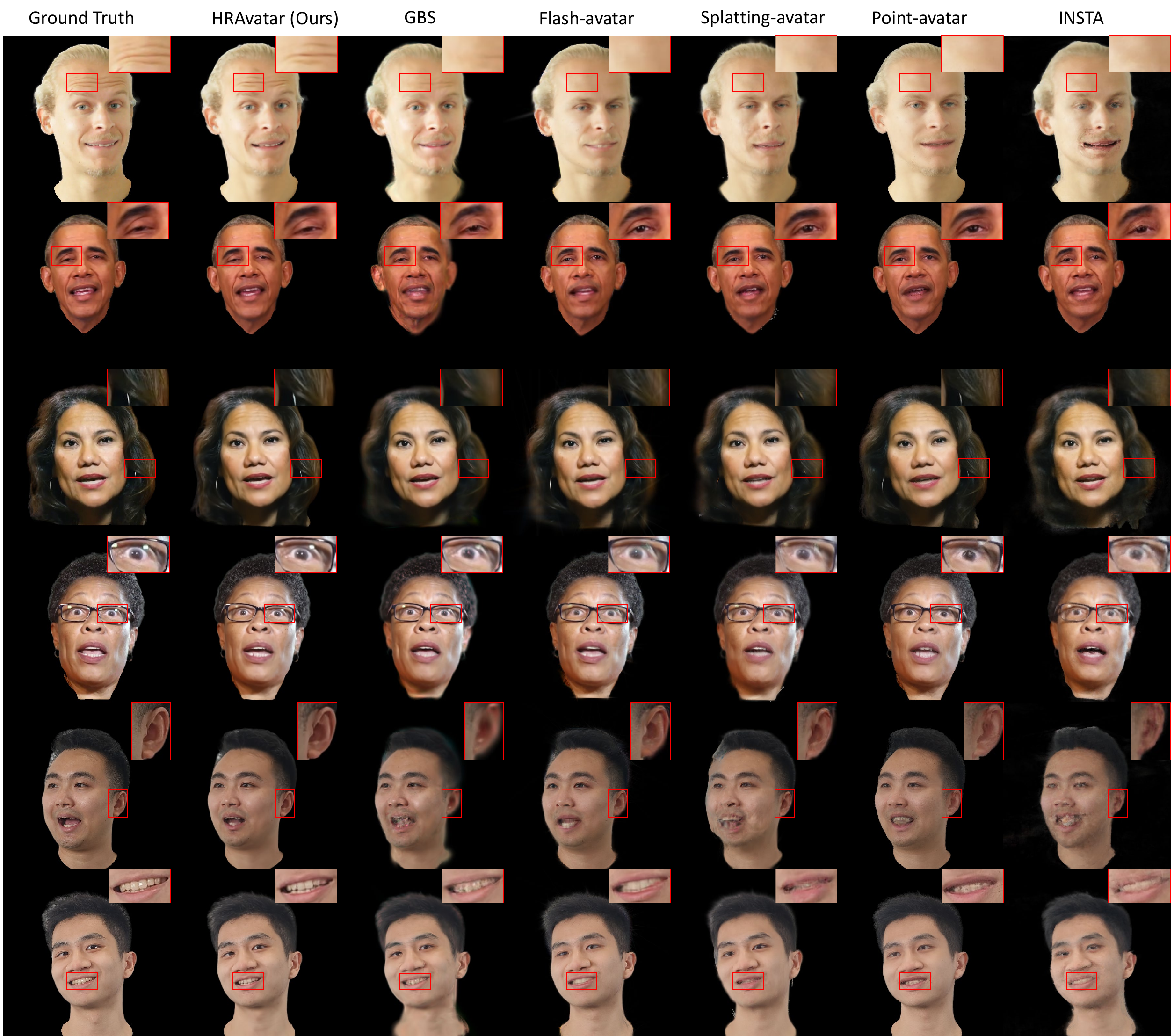}
  \caption{Qualitative comparison results on self-reenactment. Compared to others, ours captures finer texture details and renders high-fidelity images. Ours also achieves more accurate expression deformations and reconstructs better geometric details.
  }
  \label{qualitative-self-reenactment}
\end{figure*}

\begin{figure*}[t]
  \centering
  \includegraphics[width=1.0\textwidth]{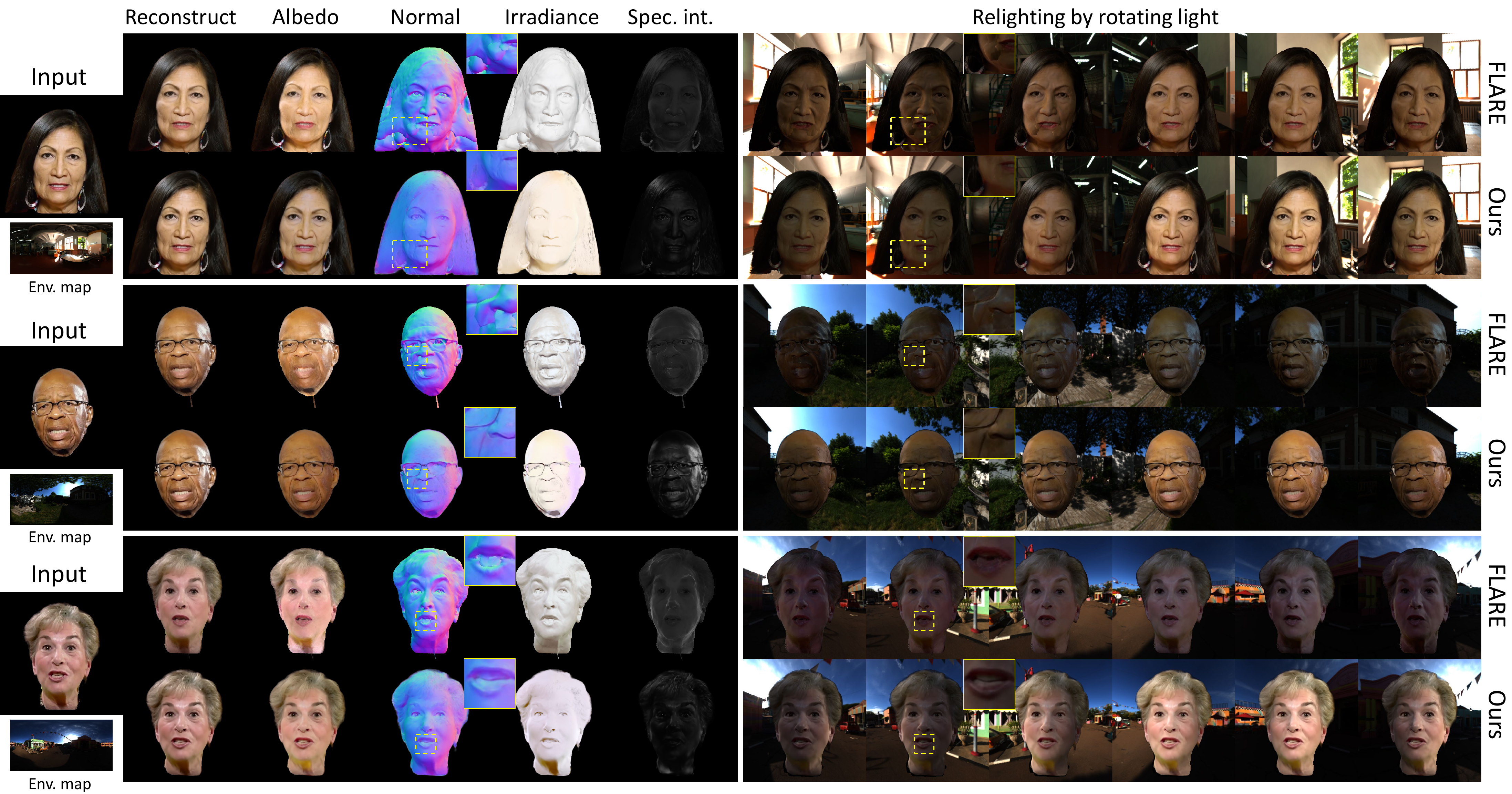}
  \caption{Visual comparison with FLARE on relighting. "Spec. int." denotes the specular intensity coefficient. FLARE exhibits some artifacts due to partially corrupted normals, while our method learns smoother normals, enabling more reasonable and consistent relighting. Notably, due to differences in pre-filtering environment maps, our method and FLARE exhibit variations in lighting brightness.
  }
  \label{flare-relight-qualitative}
\end{figure*}

\begin{figure}[t]
  \centering
  \includegraphics[width=0.48\textwidth]{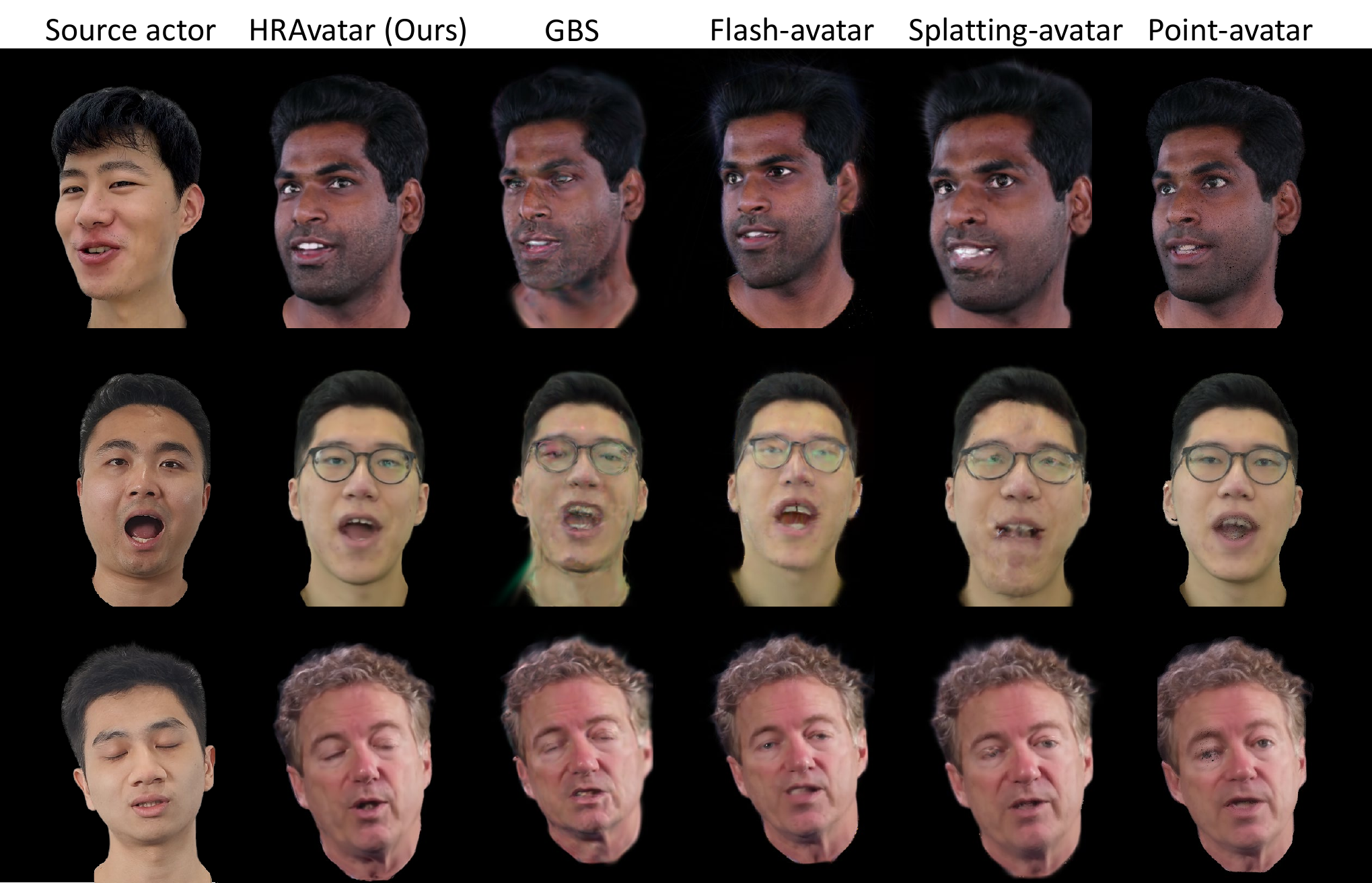}
  \caption{Visual comparison on cross-reenactment. HRAvatar accurately simulates actors' poses and expressions, preserving textures and geometric details, while others exhibit artifacts.
  }
  \label{qualitative-cross-reenactment}
\end{figure}

\begin{figure}[t]
  \centering
  \includegraphics[width=0.48\textwidth]{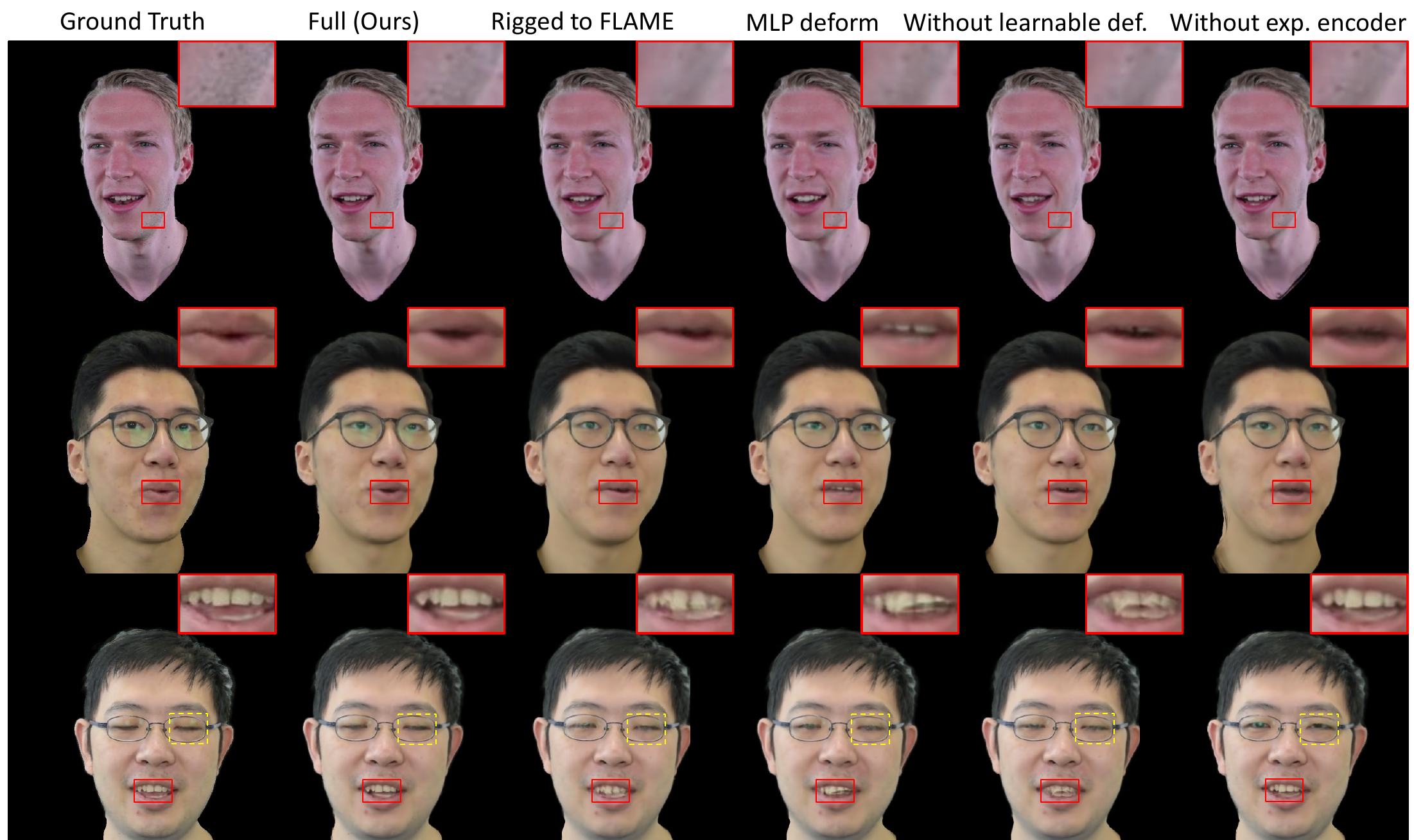}
  \caption{Qualitative results of the ablation study. Our full method renders better texture and geometry details and captures more accurate facial expressions, including mouth shapes and blinking.
  }
  \label{qualitative-quality-ablation}
\end{figure}

\begin{figure}[t]
  \centering
  \includegraphics[width=0.48\textwidth]{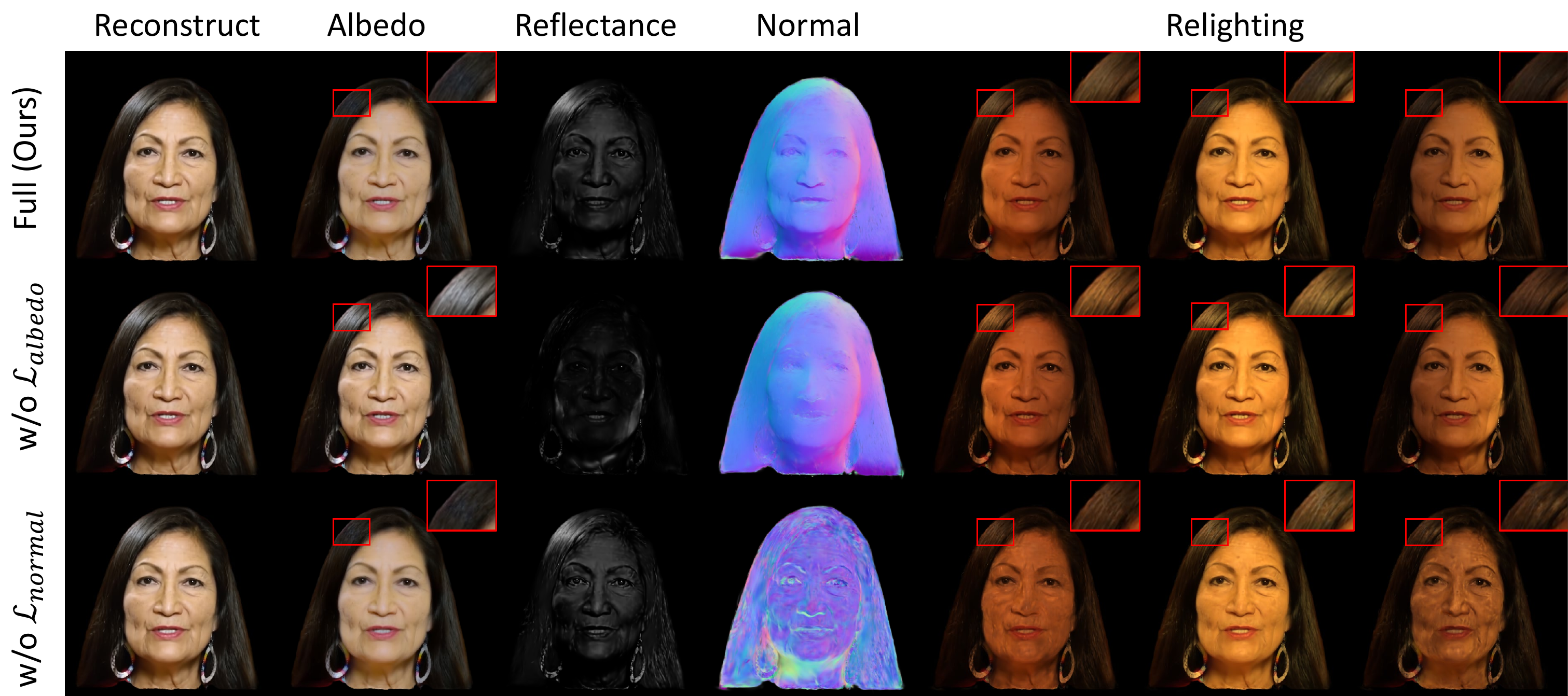}
  \caption{Ablation study for albedo and normal losses. Without $\mathcal{L}_{albedo}$, entangled attributes yield unrealistic relighting. Without $\mathcal{L}_{normal}$, chaotic normal maps cause artifacts when relighting.
  }
  \label{qualitative-relight-ablation}
\end{figure}

%% file: sec/4_expriment.tex
\section{Experiment}
\label{Experiment}
\subsection{Experimental Setup}
\label{setup}
\textbf{Implementation details.}
We build our model using PyTorch \citep{paszke2019pytorch} and train it with the Adam optimizer \citep{kingma2014adam} on a single NVIDIA 3090 GPU. Each monocular head video is trained for 15 epochs. All videos are cropped and resized to a resolution of $512\times512$. We run matting (\eg\cite{lin2022robust,chen2022robust}) to extract the foreground, setting the background to black. Moreover, we follow \citet{zheng2022avatar} to pre-track FLAME parameters for the videos. For our encoder $\mathcal{E}$, we utilize the pre-trained weight from SMIRK \citep{retsinas20243d}.

\noindent\textbf{Dataset.}
We evaluate different methods on 10 subjects from the INSTA dataset \citep{zielonka2023instant}, which provides pre-cropped and segmented images. Following INSTA, we use the last 350 frames of each video as the test set for self-reenactment evaluation. For a more robust assessment, we include 8 subjects from the HDTF dataset \citep{zhang2021flow}, which is collected from the internet. We also include 5 self-captured subjects using a mobile phone. For these two datasets, the last 500 frames are used as the test set. All methods adopt the same cropped and segmented process.

\noindent\textbf{Baseline and metrics.}
We compare our method against several SOTA methods: Point-avatar \citep{zheng2023pointavatar}, INSTA \citep{zielonka2023instant}, Splatting-avatar \citep{shao2024splattingavatar}, Flash-avatar \citep{xiang2024flashavatar}, and 3D Gaussian Blendshapes (GBS) \citep{ma20243d}, as well as FLARE \cite{bharadwaj2023flare} for relighting. For each method, we use the official code to generate the results. Note that we disable the post-training optimization of test images' parameters in Point-avatar to ensure fairness. We use PSNR, $\text{MAE}^*$ ($\text{MAE}\times10^2$), SSIM, and LPIPS \citep{zhang2018unreasonable} to evaluate the image quality.
\subsection{Evaluation}
\noindent\textbf{Quantitative results.}
We evaluate all methods for self-reenactment, as shown in \cref{self-reenactment-table}. Our method outperforms others across all four metrics, especially in LPIPS. This highlights that our method reconstructs more detailed and high-quality animatable avatars, with the improved LPIPS score suggesting sharper images. Moreover, we test HRAvatar's rendering speed for animation and relighting, achieving about \textbf{155 FPS}. Further details are in the supplementary material.

\noindent\textbf{Qualitative results.} The visual comparison of our method with baseline methods on self-reenactment is shown in \cref{qualitative-self-reenactment}. INSTA and Splatting-avatar often struggle with challenging poses, resulting in significant artifacts. Point-avatar maintains decent rendering in such poses but suffers from point artifacts and lacks detail in the mouth. Flash-avatar shows improvements but still loses some fine textures and has expression inaccuracies. GBS achieves relatively accurate facial expressions in normal poses but introduces blurring around edges, like the ears, hair, and neck. In contrast, our method accurately restores fine textures, such as hair and eye luster, while preserving precise geometric details like ears and teeth. Ours handles wrinkles and blinking more effectively due to the flexible deformation model and accurate tracking.

We qualitatively compare the visual differences in relighting between FLARE and our method. As shown in \cref{flare-relight-qualitative}, FLARE incorrectly reconstructs some of the subject's geometric normals, causing blocky artifacts during relighting. In contrast, our method learns smoother normals, leading to more consistent and realistic lighting effects. Additional comparisons with FLARE are provided in the supplementary material.

We also present cross-reenactment visual comparisons. As shown in \cref{qualitative-cross-reenactment}, our method better retains the source actor's expressions and preserves original head details, even in challenging poses and expressions, while other methods exhibit blurring and artifacts. It's worth noting that Flash-avatar and GBS treat head poses as camera poses, which may cause minor scale discrepancies, resulting in variations in the size and positioning of rendered avatars.

Additionally, the supplementary material includes more relighting results under rotating environment maps, as well as material editing and novel view synthesis.

\subsection{Ablation Studies}
The quantitative results of the ablation study on self-reenactment are summarized in \cref{ablation-table}, with qualitative results in \cref{qualitative-quality-ablation} and \cref{qualitative-relight-ablation}, validating the effectiveness of each component.

\noindent\textbf{Rigged to FLAME.} We replace HRAvatar’s learnable blendshapes and LBS with the deformation method from \citet{qian2024gaussianavatars}, which rigs Gaussian points to the FLAME mesh. The results in \cref{ablation-table} and \cref{qualitative-quality-ablation} demonstrate that our model improves on metrics and achieves more accurate texture and tooth details.

\noindent\textbf{MLP deform.} To validate the superiority of independently learning per-point blendshapes basis and blend weights, we follow Point-avatar \cite{zheng2023pointavatar} and use a shared MLP to predict them for each point. The results highlight the advantages of our learning strategy.

\noindent\textbf{Without learnable deform.} We set the blendshapes basis and blend weights as non-learnable to assess the importance of adapting to individual deformations. This leads to reduced geometry and texture quality.

\noindent\textbf{Without exp. encoder.} To verify the expression encoder's effectiveness in extracting expression parameters, we use pre-tracked parameters instead. Results indicate our method better restores facial expressions, including mouth shapes and blinking, and improves performance metrics.

\noindent\textbf{Without PBS.} This means using the standard 3DGS appearance model instead of our shading model. While the fitting-based method of 3DGS performs well due to more learnable parameters and flexibility, our method achieves comparable results while enabling realistic relighting.  

\noindent\textbf{Without $\mathcal{L}_{normal}$.} As shown in \cref{qualitative-relight-ablation}, removing normal consistency loss results in chaotic normal maps, causing blocky artifacts during relighting.

\noindent\textbf{Without $\mathcal{L}_{albedo}$.}  Without the albedo prior loss, appearance attributes become entangled, causing incorrect coupling of local highlights with albedo. This results in unrealistic relighting effects, with highlights appearing in areas without actual lighting, as shown in \cref{qualitative-relight-ablation}.

%% file: sec/5_discussion.tex
\section{Discussion}
\label{conclusion}
\textbf{Conclusion.} In this paper, we introduce HRAvatar, a novel method for high-fidelity, relightable 3D head avatar reconstruction from monocular video. To address errors incorporated from inaccurate facial expression tracking, we train an encoder in an end-to-end manner to extract more precise parameters. We model individual-specific deformations using learnable blendshapes and linear blend skinning for flexible Gaussian point deformation. By employing physically-based shading for appearance modeling, our method enables realistic relighting. Experimental results show that HRAvatar achieves state-of-the-art quality and real-time realistic relighting effects.

\noindent\textbf{Limitation.} While our method models effectively individual deformations well, it remains constrained by FLAME's priors when training data is insufficient, affecting control over elements like hair or accessories. Due to 3DGS’s strong texture representation and the limitations of existing albedo estimation models, some shadows or wrinkles may still be mis-coupled into albedo or reflectance, leading to shortcomings in relighting, particularly for specular reflections or shadows.
Besides, reconstructing the full head from a monocular video is infeasible for our method with unknown camera poses, even if the back of the head is visible. This is because monocular pose estimation relies on facial key points, which become unreliable when the yaw angle approaches 90 degrees. 

%% file: sec/X_suppl.tex
\clearpage
\appendix
\maketitlesupplementary

\newcommand{\cmark}{\textcolor{green!95!black}{\large\ding{51}}}
\newcommand{\xmark}{\textcolor{black!95!black}{\large\ding{55}}}
\newcommand{\approxmark}{\textcolor{black!95!black}{\large$\sim$}}
\definecolor{VChartColor1}{RGB}{134, 142, 255} 
\definecolor{VChartColor2}{RGB}{255, 255, 255} 
\newlength\VChartMax
\setlength\VChartMax{5em}
\newcommand*\VChart[3]{~\rlap{\textcolor{VChartColor2}{\rule{1\VChartMax}{1ex}}}\rlap{\textcolor{VChartColor2}{\rule{#3\VChartMax}{1ex}}}\rlap{\textcolor{VChartColor1}{\rule{#2\VChartMax}{1ex}}}\hphantom{\rule{1\VChartMax}{1ex}}~~~#1}

\section*{Overview}
\noindent This supplementary material presents more details and additional results not included in the main paper due to page limitation. The list of items included are:
\begin{itemize}
\item A brief description of the video results in \cref{videodemo}.
\item More model implementation details in \cref{more_implementation_details}.
\item Additional comparison with FLARE and ablation study in \cref{FurtherExperiments}.
\item Application results for novel view synthesis and material editing in \cref{AdditionalApplications}.
\item Further discussion on method differences, limitations, and ethical considerations in \cref{MoreDiscussion}.

\end{itemize}

\section{Video Demo}
\label{videodemo}
We strongly encourage readers to watch the video provided in the \href{https://eastbeanzhang.github.io/HRAvatar/}{\textcolor{magenta}{project page}}. It showcases the self-reenactment animation of avatars reconstructed by HRAvatar and includes novel view renderings.  The video also illustrates the visual results of relighting the avatars under various rotating environment maps and the ability to perform simple material editing to enhance specular reflections. Furthermore, we provide visual comparisons of HRAvatar with two advanced methods, GBS~\cite{ma20243d} and Flash-avatar~\cite{xiang2024flashavatar}, in self-reenactment, cross-reenactment, and novel view synthesis. A relighting comparison with FLARE~\cite{bharadwaj2023flare} is also included. Overall, the video highlights our method's capability to create fine-grained avatars with excellent expressiveness and realistic lighting effects in diverse environments.

\section{More Implementation Details}
\label{more_implementation_details}

\subsection{Preliminary}
3D Gaussian Splatting \citep{kerbl20233d} represents 3D scene with explicit Gaussian points, each point $G$ is defined by its position (center) $X$, rotation $r$, scaling $s$, opacity $\alpha$ and color $c$. During rendering, each Gaussian point affects nearby pixels anisotropically using a Gaussian function $\mathcal{G}$:
\begin{equation}
    \mathcal{G}(x, \mu',\Sigma_{2D})=e^{-\frac{1}{2}(x-\mu')^{\top}\Sigma_{2D}^{-1}(x-\mu') },
    \label{eq1}
\end{equation}
where $\mu'$ is the projected mean of $X$ on the image plane. Given the viewing transformation $W$, the 2D covariance matrix $\Sigma_{2D}$ is derived from the 3D covariance matrix:
\begin{equation}
    \Sigma_{2D}=JW\Sigma W^{\top}J^{\top} , \: \Sigma=RSS^{\top}R^{\top} .
    \label{eq2}
\end{equation}
$J$ is the Jacobian of the affine approximation of the projective transformation. To ensure the covariance matrix $\Sigma$ remains positive semi-definite during optimization, it is decomposed into a scaling matrix $S$ and a rotation matrix $R$, as \cref{eq2}. The scaling matrix $S$ and rotation matrix $R$ are represented by a 3D vector $s$ and a quaternion $r$, respectively. The color $c$ is modeled by a third-order spherical harmonic coefficient for view-dependent effects. During splatting, the image space is divided into multiple $16\times16$ tiles, and pixel colors are computed with alpha blending:
\begin{equation}
    \mathcal{C}(x_p)=\sum_{i\in G_{x_p}}c_i\sigma_i\prod_{j=1}^{i-1}(1-\sigma_j), \: \sigma_i=\mathcal{G}(x_p, \mu_i',\Sigma_{2D,i})\alpha_i,
    \label{eq3}
\end{equation}
where, $x_p$ represents the pixel position, and $G_{x_p}$ denotes the sorted Gaussian points associated with pixel $x_p$. Additionally, a strategy is proposed to adjust the number of Gaussian points through densification and pruning.

\begin{table}[tpb]
  \begin{center}
  \resizebox{\linewidth}{!}{
  \begin{tabular}{cccc}
    \toprule
    \  & {Rendering Quality} & {Relighting} & {Rendering speed} \\ 
    \midrule
    {Point-Avatar}~\cite{zheng2023pointavatar} & \VChart{0.646}{0.4612}{1} & Limited & $ \approx 6$ FPS \\
    {INSTA}~\citep{zielonka2023instant} & \VChart{0.764}{0.5457}{1} & \xmark & $ \approx 1$ FPS \\
    {FLARE}~\cite{bharadwaj2023flare} & \VChart{0.698}{0.4985}{1} & \cmark & $ \approx 35$ FPS \\
    {Splatting-avatar}~\cite{shao2024splattingavatar} & \VChart{0.834}{0.5955}{1} & \xmark & $>120$ FPS \\
    {Flash-avatar}~\cite{xiang2024flashavatar} & \VChart{0.883}{0.6305}{1} &\xmark & $>120$ FPS \\
    {GBS}~\cite{ma20243d} & \VChart{0.980}{0.7001}{1} & \xmark & $>120$ FPS \\
    {HRAvatar (Ours) } & \VChart{1.184}{0.8457}{1} & \cmark & $>120$ FPS \\
    \bottomrule
  \end{tabular}
  }
    \caption{Key aspects of our method compared to previous works. The rendering quality shows the inverse of the MAE metric on the INSTA dataset, with longer bars representing better performance. 'Limited' indicates that the Point-Avatar method has limited flexibility in handling relighting.}
  \label{key-aspects-tab}
  \end{center}
\end{table}

\subsection{Training Details}
In the first 1500 iterations, we take the albedo map as the rendered image to learn the head's albedo properties initially. Afterward, we switch to shaded image to learn other attributes. Each Gaussian point's roughness, Fresnel base reflectance, and albedo attributes are initialized to 0.9, 0.04, and 0.5, respectively. While we generally follow 3DGS hyperparameters, we make some adjustments. During training, point densification starts at iteration 1000 and ends at 500 iterations before training completes, with a densification interval of 500 iterations. The gradient threshold is increased to $3\times10^{-4}$ to avoid excessive point growth. During training, opacity is reset below the pruning threshold to eliminate more redundant points. The learning rates for the Gaussian point positions, appearance attributes, and environment map gradually decrease as training progresses, while the expression encoder learning rate is set to $5\times10^{-5}$. Training a video with 2400 frames takes about one hour. 

When using albedo prior to supervision, we apply it every 3 frames due to the time-consuming process of extracting pseudo-ground-truth albedo during preprocessing. Additionally, since the lighting in the INSTA and self-captured datasets is relatively uniform, we only apply albedo prior supervision during training on the HDTF dataset. Furthermore, for subjects in the HDTF dataset, we set a higher upper bound for reflectance ($\tau_{max}^{f_0}$) to account for the specific lighting conditions.

\begin{table*}[htbp]
\resizebox{\linewidth}{!}{
\begin{tabular}{ccccccccccccc}
     \toprule
       \multirow{2}{*}{Method}  & \multicolumn{4}{c}{INSTA dataset} & \multicolumn{4}{c}{HDTF dataset}  &  \multicolumn{4}{c}{self-captured dataset} \\
      \cline{2-5}    \cline{6-9}     \cline{10-13} 
     & PSNR$\uparrow$ & $\text{MAE}^*$$\downarrow$ & SSIM$\uparrow$ & LPIPS$\downarrow$ & PSNR$\uparrow$ & $\text{MAE}^*$$\downarrow$ & SSIM$\uparrow$ & LPIPS$\downarrow$ & PSNR$\uparrow$ & $\text{MAE}^*$$\downarrow$ & SSIM$\uparrow$ & LPIPS$\downarrow$ \\
    \midrule
   FLARE   & 26.80 &1.433 & 0.9063 & 0.0816 & 25.55 &2.193 & 0.8479 & 0.1183 & 25.82 &1.715 & 0.8576 & 0.1230 \\
   HRAvatar (Ours)   & \bf30.36 & \bf0.845 & \bf0.9482 & \bf0.0569 & \bf28.55 & \bf1.373 & \bf0.9089 & \bf0.0825 & \bf28.97 & \bf1.123 & \bf0.9054 & \bf0.1059 \\
   \bottomrule
\end{tabular}
}
\caption{Average quantitative results on the INSTA, HDTF, and self-captured datasets. Our method outperforms FLARE in PSNR, $\text{MAE}^*$ ($\text{MAE}\times10^2$), SSIM, and LPIPS metrics.}
\label{flare-self-reenactment-table}
\end{table*}

\begin{table}[tb]
\centering
\resizebox{1.0\linewidth}{!}{
\begin{tabular}{cc|c}
     \toprule
       & albedo (LMSE $\downarrow$) & normal (cosine similarity $\uparrow$)  \\
    \midrule
    FLARE  &  0.0665 & 0.8424  \\ 
    Ours & \bf0.0557 & \bf0.9093  \\ 
   \bottomrule
\end{tabular}
}
\caption{Albedo and normal evaluation on the HDTF Dataset.}
\label{tb-intrinsic}
\end{table}

\subsection{Model Details}
The shape and expression basis in FLAME are derived through PCA, with higher dimensions having a small effect on deformation. To avoid unnecessary computations, we use only the first 100 shape parameters and 50 expression parameters, i.e., $|\beta|=100$ and $|\psi|=50$.
Since FLAME lacks an interior mesh for the mouth, we follow \citet{qian2024gaussianavatars} by adding a mesh for the teeth, where the upper and lower teeth move according to the neck and jaw joints, respectively. Additionally, we add extra mesh behind the teeth to provide a reasonable initialization for the rest of the mouth interior.

During shading, normal and reflection vectors sample lighting from the irradiance and pre-filtered environment maps. Since both maps must be backpropagated and mipmaps reconstructed in each training iteration, the computation increases with resolution.  To maintain efficient training, we set the irradiance map $I_{irr}$ resolution to $16\times16$ and the pre-filtered environment map $I_{env}$ to $32\times32$  with 3 mipmap levels.

\subsection{BRDF Reflection Model.}
For physical-based shading, we use the Disney model~\citep{burley2012physically} to describe light interactions with geometry and materials, a method commonly employed in real-time rendering. This model breaks reflection into two components: Lambertian diffuse reflection and specular reflection:
\begin{equation}
\begin{split}
     L_o(X,\omega_o)=L_d+L_s=\int_{\Omega}{\frac{a}{\pi}}L_i(X,\omega_i)n\cdot\omega_i d\omega_i\\
+\int_{\Omega}{\frac{\mathcal{DFH}}{4(n\cdot\omega_o)(n\cdot\omega_i)}}L_i(X,\omega_i)n\cdot\omega_i d\omega_i ,
\end{split}
\label{eq16}
\end{equation}
where $L_i$ and $L_o$ denote the radiance for the incoming direction $\omega_i$ and outgoing direction $\omega_o$, respectively with $n$ as the normal. The Lambertian term models diffuse reflection, independent of viewing direction, allowing us to precompute and store this part in an irradiance map. The specular reflection term models appearance based on viewing angle, with $\mathcal{D}$, $\mathcal{F}$, and $\mathcal{H}$ representing the normal distribution, Fresnel equation, and geometric function. We use the SplitSum approximation to simplify the BRDF integral into two parts:
\begin{equation}
\begin{split}
L_s&\approx I_{env}\cdot I_{BRDF}=(\frac{1}{\mathrm{Z}}\sum_{z=1}^{\mathrm{Z}}L_i(\omega_z))
\\&\cdot(\frac{1}{\mathrm{Z}}\sum_{z=1}^{\mathrm{Z}}\frac{\mathcal{DFH}\cdot n\cdot\omega_z}{4(n\cdot\omega_o)(n\cdot\omega_z)pdf(\omega_z,\omega_o)})
~.
\end{split}
\label{eq17}
\end{equation}
Here, $pdf(\omega_m,\omega_o)$ is the probability density function related to $\mathcal{D}$. Both components are precomputed and stored: $I_{env}$ as a multi-resolution mipmap for different roughness levels and $I_{BRDF}$, as a lookup table (LUT) based on roughness and the dot product of the normal and observation direction, $n\cdot\omega_o$.

\begin{figure}[t]
  \centering
  \includegraphics[width=0.48\textwidth]{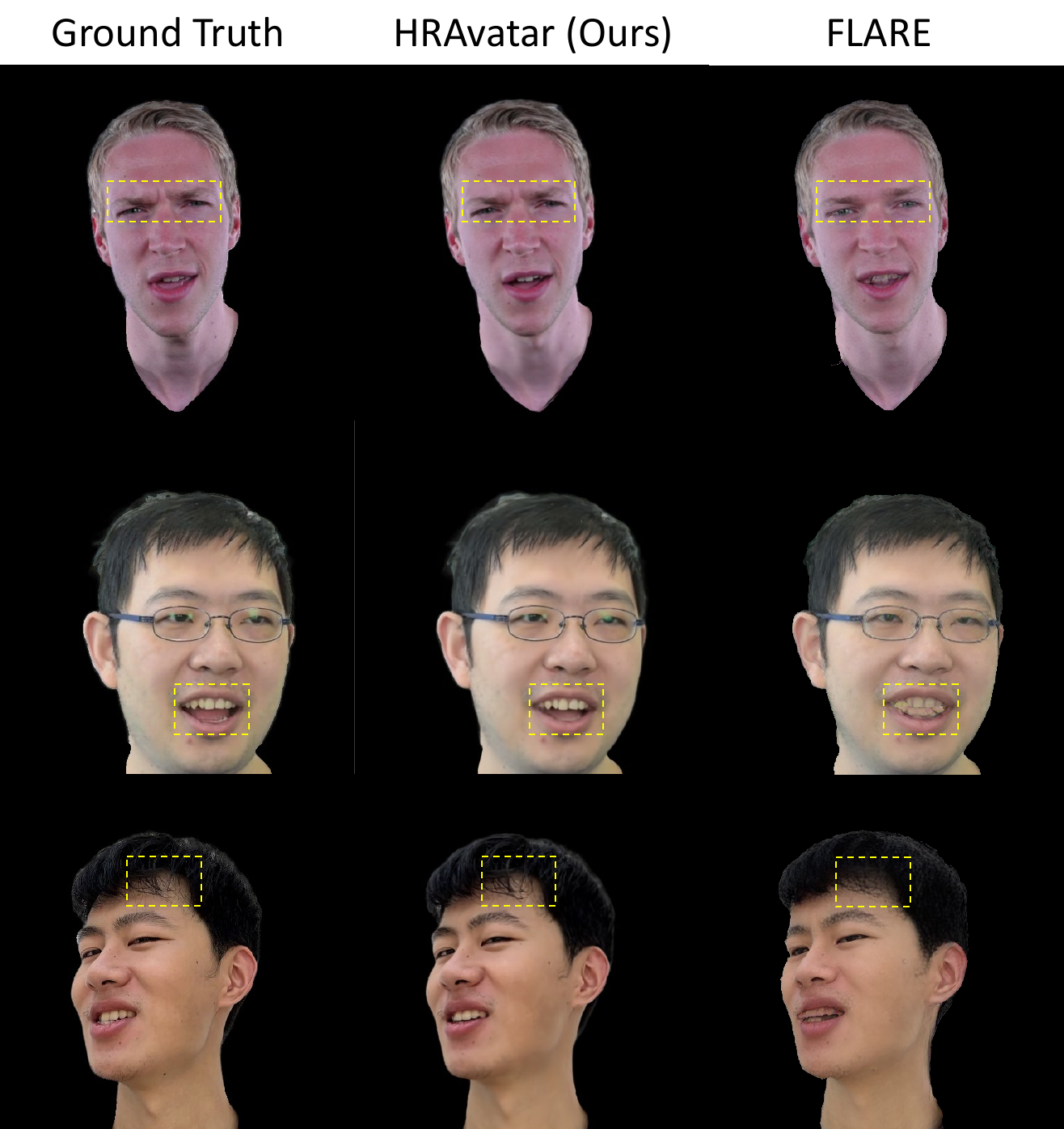}
  \caption{Visual comparison with FLARE on self-reenactment. Our method captures facial expression details more effectively and reconstructs the teeth geometry and hair texture more accurately.
  }
  \label{flare-self-reenactment-qualitative}
\end{figure}

\section{Further Experiments}
\label{FurtherExperiments}
\subsection{Rendering Speed}
Despite the additional computational load introduced by the deformation and appearance models, our method still achieves real-time rendering speeds. To provide a reference, we test the rendering speed on the INSTA dataset using a single NVIDIA 3090 GPU. Each trained avatar contains about 75K Gaussian points. We set the rendering resolution to $512\times512$ and render 500 images to calculate the average speed. HRAvatar achieves an average speed of about \textbf{155 FPS}, with the encoder extracting parameters at about 179 FPS. Similarly, when \textbf{relighting} with a new environment map, we measured a rendering speed of approximately \textbf{155 FPS} under the same setup, ensuring real-time performance.

\subsection{Comparison with FLARE}
Since both FLARE~\cite{bharadwaj2023flare} and our method can perform monocular 3D head reconstruction and relighting, we conduct a further comparison. 

\noindent\textbf{Self-reenactment.} The experimental setup is the same as in the main paper, with quantitative results shown in \cref{flare-self-reenactment-table} and qualitative results in \cref{flare-self-reenactment-qualitative}. Our method outperforms FLARE in both metrics and visual quality, better capturing details of facial expressions, hair textures, and internal mouth features such as teeth.

 \noindent\textbf{Speed.} Under the same setup, we test FLARE's average rendering speed on the INSTA dataset, which is approximately 35 FPS. In contrast, our method achieves a rendering speed of about \textbf{4.5×} higher.

\noindent\textbf{Disentanglement and geometric.} Directly evaluating material disentanglement is challenging due to the scarcity of publicly available real or synthetic face video datasets. As an alternative, we employ SwitchLight \cite{kim2024switchlight} to extract image albedo as pseudo-ground truth for evaluation. We compare against FLARE using LMSE (Local Mean Squared Error) \cite{grosse2009lmse} as the evaluation metric. Results are in \cref{tb-intrinsic}. Roughness and reflectance are excluded due to varying definitions and usage across shading models.

Normals are commonly used to assess reconstructed 3D geometry. To quantify this, since we lack ground truth normals, we use the SOTA single-image geometry estimation method GeoWizard \cite{fu2024geowizard}, to estimate normals from the images as pseudo-ground truth. We use the cosine similarity of normals as the evaluation metric, as shown in \cref{tb-intrinsic}.

The qualitative comparison of normals and decoupling results is shown in \cref{flare-relight-qualitative} of the main paper.

\subsection{Ablation Of Jaw Pose Regularization Loss}%
\begin{figure}[t]
  \centering
  \includegraphics[width=0.48\textwidth]{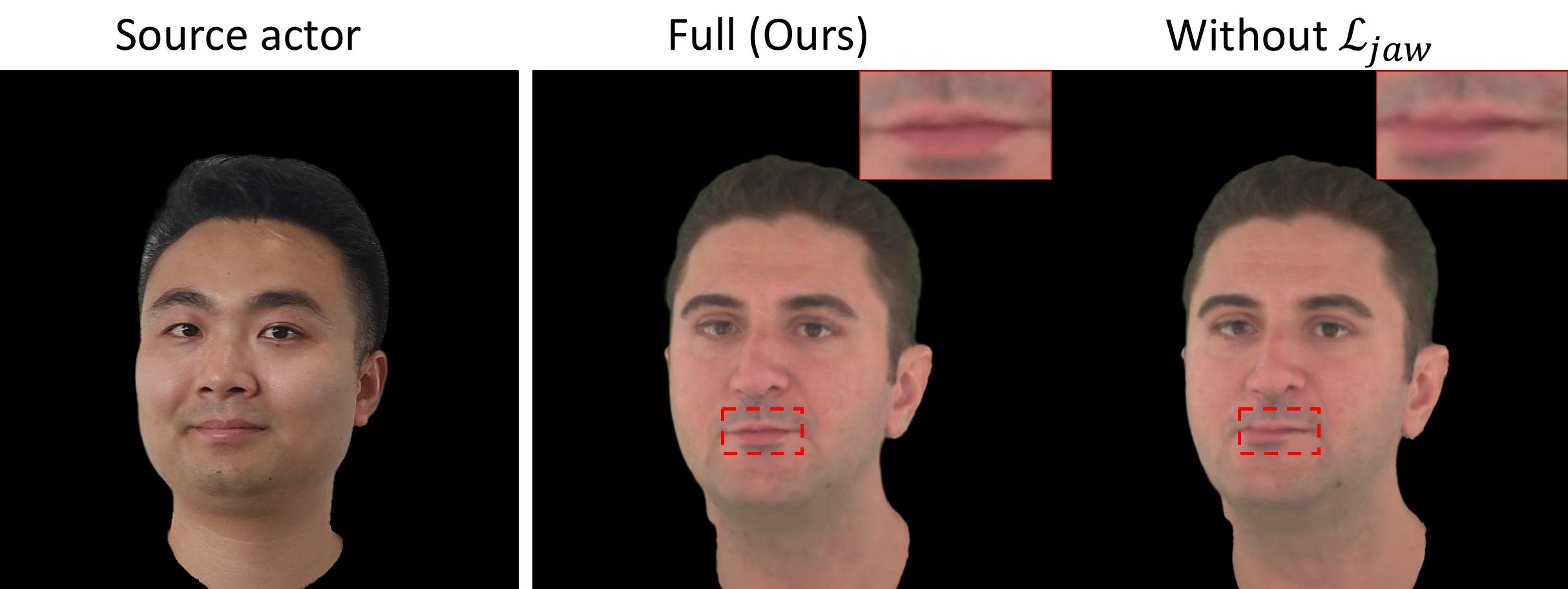}
  \caption{Ablation result on $\mathcal{L}_{jaw}$. Without the jaw pose regularization loss, the avatar exhibits mouth distortion during cross-reenactment.
  }
  \label{ablation-wo-reg-jaw}
\end{figure}

Without the jaw pose regularization loss, $\mathcal{L}_{jaw}$, the trained encoder may extract jaw poses that deviate from the normal distribution. This can lead to incorrect mouth motion during cross-reenactment. As shown in \cref{ablation-wo-reg-jaw}, removing $\mathcal{L}_{jaw}$ results in mouth distortion, while including this loss effectively prevents the issue.

\subsection{Complete Quantitative Results}
We present the complete quantitative results of self-reenactment for each subject on the INSTA, HDTF, and self-captured datasets in \cref{self-reenactment-insta-full—table} and \cref{self-reenactment-rest-full—table}. As shown, HRAvatar achieves superior performance for most subjects, demonstrating the robustness of our method.

\begin{table*}[htbp]
\begin{center}
\resizebox{0.7\linewidth}{!}{
\begin{tabular}{c|ccccccccccc}
     \toprule
     \multicolumn{2}{c}{} & \multicolumn{10}{c}{INSTA dataset}   \\
      \cline{3-12} 
    \multicolumn{2}{c}{} & bala & biden & justin  & malte\_1 &marcel & nf\_01 & nf\_03 & obama & person0004 & wojtek\_1 \\
    \midrule  
    \multirow{6}{*}{PSNR$\uparrow$} 
   &INSTA   & 29.53 & 29.92 & 31.66 & 27.44 & 22.99 & 26.45 & 28.31 & 31.21 & 25.44 &31.36 \\
   &Point-avatar & 27.88 & 27.64 & 30.40 & 24.98 & 24.66 & 25.25 & 26.60 & 28.83 & 23.29 &28.82 \\
   &FLARE & 27.20 & 28.55 & 29.10 & 25.93 & 22.50 & 25.97 & 26.71 & 28.67 & 25.53 &27.84 \\
   &Splatting-avatar& 32.14 & 30.42 & 30.93 & 27.66 & 24.34 & 27.08 & 27.85 & 30.64 & \underline{26.49} &29.54 \\
   &Flash-avatar& 30.27 & 31.25 & 32.16 & 27.45 & 24.85 & \underline{28.02} & \underline{28.28} & \underline{31.46} & 25.49 &\underline{32.03} \\
   &GBS & \underline{32.47} & \bf32.23 & \underline{33.10} & \underline{28.23} & \underline{26.11} & 27.59 & 28.12 & 31.35 & 25.16 &\bf32.05 \\
   &HRAvatar (Ours)   & \bf33.10 & \underline{31.70} & \bf33.29 & \bf29.28 & \bf26.58 & \bf28.95 & \bf29.68 & \bf33.24 & \bf26.54 & 31.26 \\
       \midrule  
    \multirow{6}{*}{$\mathrm{MAE^*}$$\downarrow$} 
   &INSTA   & 1.154 & 0.849 & 0.642 & 1.160 & 2.996 & 1.705 & 1.381 & 0.775 & 1.594 &0.834 \\
   &Point-avatar & 1.386 & 1.203 & 0.869 & 1.596 & 2.662 & 1.800 & 1.583 & 1.103 & 2.083 &1.042 \\
   &FLARE & 1.342 & 0.973 & 0.910 & 1.470 & 2.817 & 1.706 & 1.602 & 1.097 & 1.392 &1.020 \\
   &Splatting-avatar& 0.854 & 0.838 & 0.783 & 1.135 & 2.309 & 1.533 & 1.340 & 0.917 & \underline{1.376} &0.910 \\
   &Flash-avatar& 1.175 & 0.670 & 0.610 & 1.058 & 2.133 & 1.326 & 1.249 & 0.819 & 1.589 &0.700 \\
   &GBS & \underline{0.747} & \bf0.583 & \underline{0.520} & \underline{1.010} & \underline{1.608} & \underline{1.311} & \underline{1.162} & \underline{0.802} & 1.803 &\bf0.655 \\
   &HRAvatar (Ours)   & \bf0.657 & \underline{0.616} & \bf0.498 & \bf0.902 & \bf1.293 & \bf1.133 & \bf1.031 & \bf0.580 & \bf1.070 & \underline{0.668} \\
       \midrule  
    \multirow{6}{*}{SSIM$\uparrow$} 
   &INSTA   & 0.8896 & 0.9460 & 0.9591 & 0.9159 & 0.8736 & 0.8937 & 0.8676 & 0.9484 & 0.8478 &0.9452 \\
   &Point-avatar & 0.8658 & 0.9116 & 0.9373 & 0.8853 & 0.9063 & 0.8919 & 0.8807 & 0.9145 & 0.8576 &0.9192 \\
   &FLARE & 0.8761 & 0.9347 & 0.9363 & 0.8973 & 0.8892 & 0.9027 & 0.8841 & 0.9199 & 0.9015 &0.9216 \\
   &Splatting-avatar& 0.9272 & 0.9466 & 0.9482 & 0.9243 & 0.9041 & 0.9202 & 0.9113 & 0.9411 & \underline{0.9075} &0.9400 \\
   &Flash-avatar& 0.8494 & 0.9614 & 0.9611 & 0.9326 & 0.9086 & 0.9270 & 0.9155 & \underline{0.9493} & 0.8996 & 0.9509 \\
   &GBS & \underline{0.9390} & \bf0.9658 & {0.9690} & \underline{0.9374} & \underline{0.9217} & \underline{0.9365} & \underline{0.9271} & 0.9476 & 0.8910 &\bf0.9593 \\
   &HRAvatar (Ours)   & \bf0.9473 & \underline{0.9635} & \underline{0.9687} & \bf0.9429 & \bf0.9352 & \bf0.9398 & \bf0.9334 & \bf0.9647 & \bf0.9278 & \underline{0.9590} \\
   \midrule  
       \multirow{6}{*}{LPIPS$\downarrow$} 
   &INSTA   & 0.0992 & 0.0541 & 0.0521 & 0.0731 & 0.1351 & 0.1262 & 0.1286 & 0.0446 & 0.1453 &0.0540 \\
   &Point-avatar & 0.0829 & 0.0637 & 0.0588 & 0.0758 & 0.1247 & 0.1257 & 0.1143 & 0.0589 & 0.1637 &0.0576 \\
   &FLARE & 0.0927 & 0.0513 & 0.0582 & 0.0726 & 0.1266 & 0.1068 & 0.0971 & 0.0595 & \underline{0.0947} &0.0567 \\
   &Splatting-avatar& 0.0865 & 0.0564 & 0.0651 & 0.0749 & 0.1326 & 0.1107 & 0.0966 & 0.0545 & 0.1246 &0.0602 \\
   &Flash-avatar&0.1535 & {0.0299} & \underline{0.0378} & \underline{0.0477} & \underline{0.1069} & \underline{0.0868} & \underline{0.0760} & \underline{0.0376} & 0.1035 &\underline{0.0392} \\
   &GBS & \underline{0.0862} & 0.0433 & 0.0481 & 0.0737 & 0.1219 & 0.1076 & 0.0861 & 0.0564 & 0.1417 &0.0582 \\
   &HRAvatar (Ours)   & \bf0.0451 & \underline{0.0306} & \bf0.0367 & \bf0.0476 & \bf0.0992 & \bf0.0868 & \bf0.0649 & \bf0.0279 & \bf0.0940 & \bf0.0358 \\
   \bottomrule
\end{tabular}
}
\end{center}
\caption{Complete quantitative results of self-reenactment for each subject on the INSTA dataset. HRAvatar achieves better performance metrics in most cases. {Bold} marks the best, and \underline{underline} marks the second.}
\label{self-reenactment-insta-full—table}
\end{table*}
\begin{table*}[htbp]
\begin{center}
\resizebox{0.7\linewidth}{!}{
\begin{tabular}{c|ccccccccc|ccccc}
     \toprule
     \multicolumn{2}{c}{} & \multicolumn{8}{c}{HDTF dataset} & \multicolumn{5}{c}{self-captured dataset}  \\
      \cline{3-15} 
    \multicolumn{2}{c}{} & elijah & haaland & katie  & marcia &randpaul & schako & tom & veronica & s1 & s2  & s3 &s4 & s5  \\
    \midrule  
    \multirow{6}{*}{PSNR$\uparrow$} 
   &INSTA   & 25.00 & 24.94 & 21.36 & 24.61 & 23.50 & 26.45 & 29.16 & 26.45       & 25.88 &25.37 &29.33 &24.86 &24.086 \\
   &Point-avatar & 24.05 & 25.56 & 22.51 & 23.76 & 26.28 & 25.44 & 27.01 & 26.51       & 25.35 &27.32 &28.09 &23.56 &24.85 \\
   &FLARE & 25.05 & 25.66 & 22.10 & 23.58 & 26.98 & 25.05 & 29.45 & 26.50      & 26.26 &26.12 &28.32 &24.07 &24.32 \\
   &Splatting-avatar& 26.08 & 26.31 & 22.23 & 25.80 & 29.25 & 25.51 & 30.98 & 27.14       & 25.05 &28.20 &29.54 &25.34 &24.22 \\
   &Flash-avatar& 26.29 & 26.46 & \underline{23.39} &\underline{ 26.67} & 29.05 & \bf28.28 & \underline{31.56} & \underline{28.95}       & 26.37 &27.26 &30.59 &\bf28.01 &25.09 \\
   &GBS & \underline{26.76} & \underline{28.29} & 22.74 & 26.59 & \underline{29.20} & 27.88 & 31.54 & 29.48       &\underline{28.15} &\underline{29.50} &\bf31.64 &\underline{27.48} & \underline{26.17} \\
   &HRAvatar (Ours)   & \bf28.24 & \bf28.91 & \bf24.92 & \bf27.23 & \bf29.70 & \underline{27.95} & \bf31.75 & \bf29.71 & 
        \bf29.40 & \bf30.19 &\underline{31.40} &27.00 &\bf26.84\\
    \midrule  
    \multirow{6}{*}{$\text{MAE}^*$$\downarrow$} 
   &INSTA   & 1.835 & 2.161 & 4.179 & 2.191 & 2.602 & 1.936 & 1.272 & 2.487       & 1.877 &1.637 &1.377 &1.841 &2.807 \\
   &Point-avatar & 2.058 & 2.177& 3.493 & 2.423 & 1.746 & 2.092 & 1.683 & 2.212       & 1.852 &1.312 &1.204 &1.903 &2.210 \\
   &FLARE & 1.813 & 2.097& 3.732 & 2.580 & 1.637 & 2.207 & 1.204 & 2.277      & 1.762 &1.540 &1.209 &1.736 &2.328 \\
   &Splatting-avatar& 1.652 & 1.915 & 3.841 & 2.026 & 1.260 & 2.200 & 0.988 & 2.183       & 2.093 &1.296 &1.110 &1.565 &2.489 \\
   &Flash-avatar& 1.602 & 2.052 & \underline{2.922} &1.755 & 1.312 & 1.519 & 0.980 & 1.865       & 1.909 &1.364 &1.079 &\underline{1.251} &2.557 \\
   &GBS & \underline{1.406} & \underline{1.403} & 3.216 & \underline{1.659} & \underline{1.234} & \underline{1.452} & \underline{0.901} & \underline{1.535}       &\underline{1.379} &\underline{1.022} &0.950 &1.285 & \underline{2.018} \\
   &HRAvatar (Ours)   & \bf1.108 & \bf1.319 & \bf2.283 & \bf1.483 & \bf1.079 & {1.384} & \bf0.847 & \bf1.477 & 
        \bf1.142 & \bf0.896 &\bf0.792 &\bf1.117 &\bf1.666\\
       \midrule  
    \multirow{6}{*}{SSIM$\uparrow$} 
   &INSTA   & 0.8808 & 0.8337 & 0.7474 & 0.8290 & 0.8528 & 0.8586 & 0.9143 & 0.7700      & 0.8218 &0.8659 &0.8722 &0.8634 &0.7431 \\
   &Point-avatar & 0.8631 & 0.8275 & 0.7771 & 0.8160 & 0.8694 & 0.8578 & 0.8634 & 0.8339       & 0.8460  &0.8763 &0.8867 &0.8573 &0.8117  \\
   &FLARE & 0.8798 & 0.8426 & 0.7773 & 0.8117 & 0.8773 & 0.8517 & 0.9064 & 0.8364       & 0.8522  &0.8560 &0.8878 &0.8716 &0.8204  \\
   &Splatting-avatar& 0.8952 & 0.8562 & 0.7562 & 0.8477 & 0.9094 & 0.8586 & 0.9321 & 0.8337       & 0.8279 &0.8775 &0.9038 &0.8817 &0.8031 \\
   &Flash-avatar& 0.8898 & 0.8146 & \underline{0.8133} & 0.8636 & 0.9040 & 0.8982 & 0.9305 & 0.8170       & 0.7774 &0.8659 &0.8967 &0.8850 &0.7491 \\
   &GBS & \underline{0.9113} &\underline{ 0.8924} & 0.8068 & \underline{0.8783} & \underline{0.9110} &\underline{ 0.9091} & \underline{0.9404} & \underline{0.8826}       & \underline{0.8799} &\underline{0.9098} &\underline{0.9188} &\underline{0.9029} &\underline{0.8339} \\
   &HRAvatar (Ours)   & \bf0.9335 & \bf0.9036 & \bf0.8597 & \bf0.8961 & \bf0.9254 & \bf0.9135 & \bf0.9446 & \bf0.8951        & \bf0.9019 & \bf0.9232 &\bf0.9283 &\bf0.9142 &\bf0.8596 \\
   \midrule  
       \multirow{6}{*}{LPIPS$\downarrow$} 
   &INSTA   & 0.1005 &0.1698 & 0.2222 & 0.1586 & 0.1417 & 0.1390 & 0.0729 & 0.2415       & 0.1897  &0.1583  &0.1523  &0.1678 &0.2483\\
   &Point-avatar &0.0886 & 0.1360 & 0.1683 & 0.1200 & 0.1147 & 0.1283 & 0.0981 & 0.1686       & 0.1255 &0.0942 &0.1024 &0.1364 &0.1623 \\
   &FLARE &0.0821 & 0.1255 & 0.1589 & 0.1258 & 0.1040 & 0.1193 & 0.0748 & 0.1559       & 0.1217 &0.1014 &0.1088 &0.1331 &\underline{0.1500} \\
   &Splatting-avatar& 0.0902 & 0.1476 & 0.1982 & 0.1385 & 0.1033 & 0.1455 & 0.0664 &0.1907       &0.1773 &0.1271 &0.1194 &0.1539 &0.1972 \\
   &Flash-avatar&\underline{0.0759} & 0.1595 & \underline{0.1387} & \underline{0.0881} & \underline{0.0829} & \underline{0.1011} & \underline{0.0609} & \underline{0.1688}        & 0.2346 &\underline{0.0736} &\bf0.0901 &\bf0.109 &0.2208  \\
   &GBS & 0.0875 &\underline{ 0.1515} & 0.1899 & 0.1289 & 0.1113 & 0.1160 & 0.0679 & 0.1850        & \underline{0.1696} &0.1198 &0.1305 &0.1599 &0.2004  \\
   &HRAvatar (Ours)   & \bf0.0504 & \bf0.0929 & \bf0.1208 & \bf0.0723 & \bf0.0683 & \bf0.0846 & \bf0.0485 & \bf0.12228         & \bf0.1063 &\bf0.0662 &\underline{0.0939} &\underline{0.1153} &\bf0.1478 \\
   \bottomrule
\end{tabular}
}
\end{center}
\caption{Complete quantitative results of self-reenactment for each subject on the HDTF and self-captured dataset. HRAvatar achieves better performance metrics in most cases.}
\label{self-reenactment-rest-full—table}
\end{table*}

\begin{figure*}[t]
  \centering
  \includegraphics[width=1.0\textwidth]{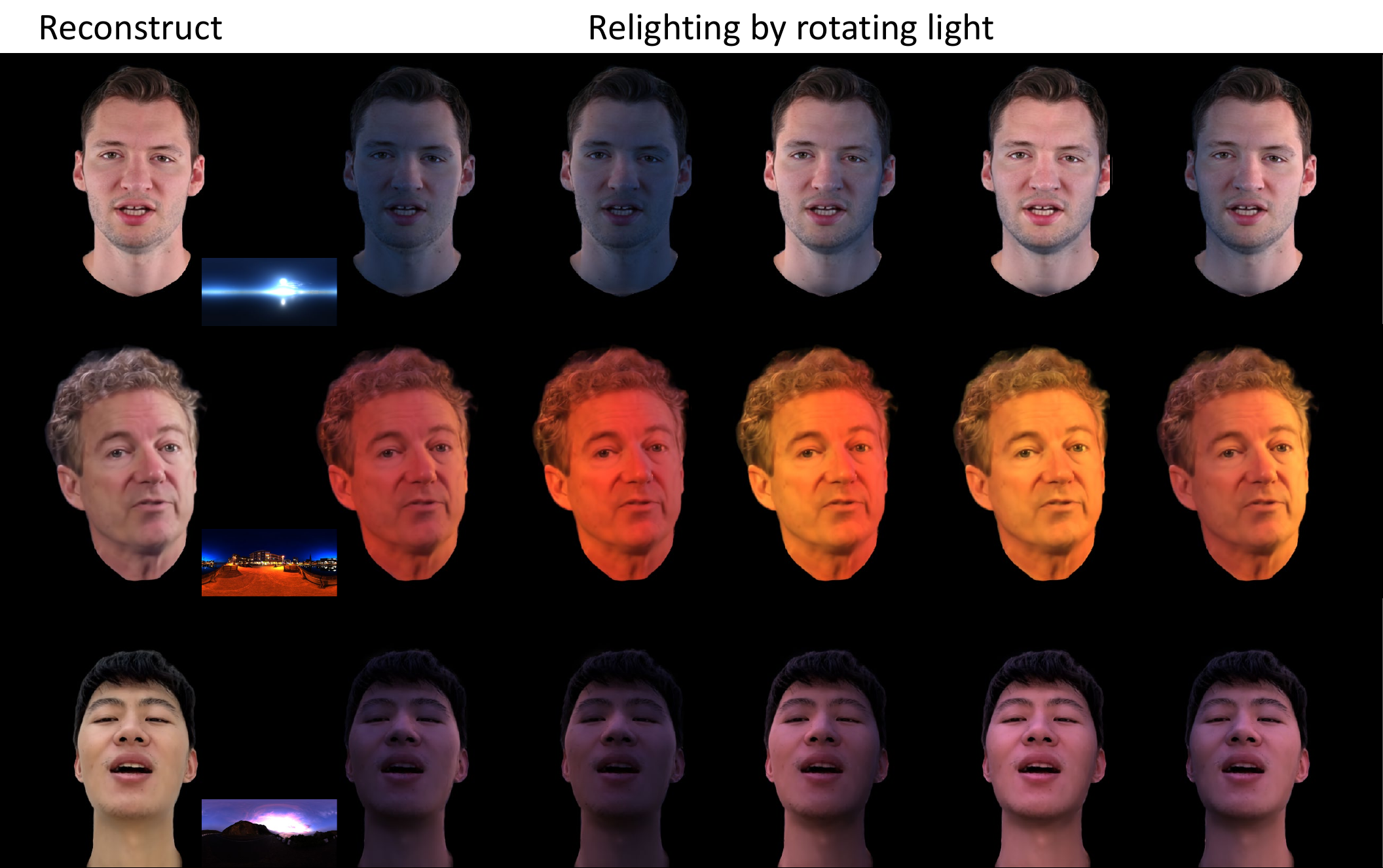}
  \caption{Relighting visual results. For each environment map, we rotate the lighting to illuminate the head from different directions.
  }
  \label{show-relight}
\end{figure*}

\begin{figure*}[t]
  \centering
  \includegraphics[width=\textwidth]{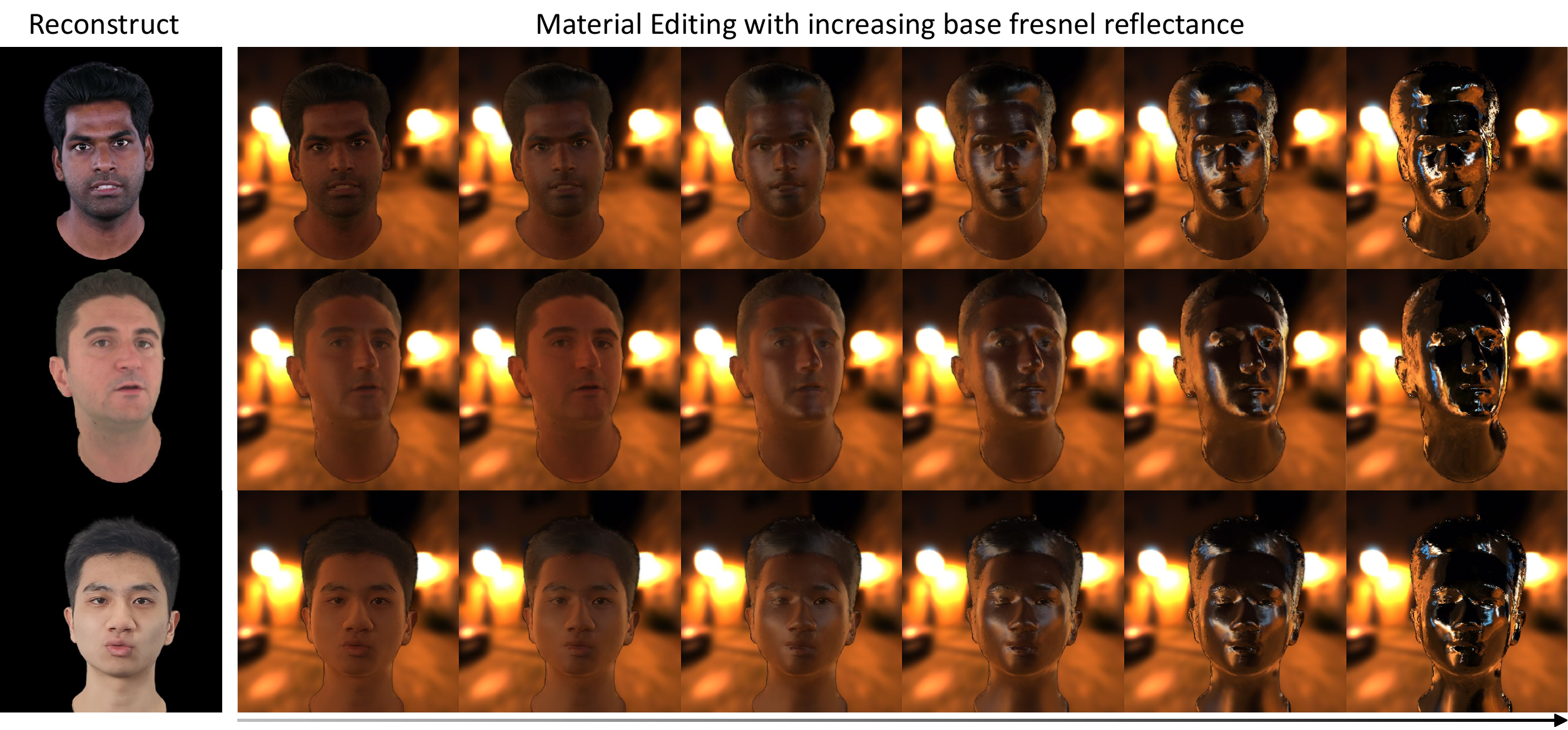}
  \caption{Visual results of material editing. We gradually increase the avatar's base Fresnel reflectance under new environment lighting, enhancing specular reflections. The results align with intuitive expectations, validating the effectiveness of our shading model.
  }
  \label{show-material-editing}
\end{figure*}

\begin{figure*}[t]
  \centering
  \includegraphics[width=\textwidth]{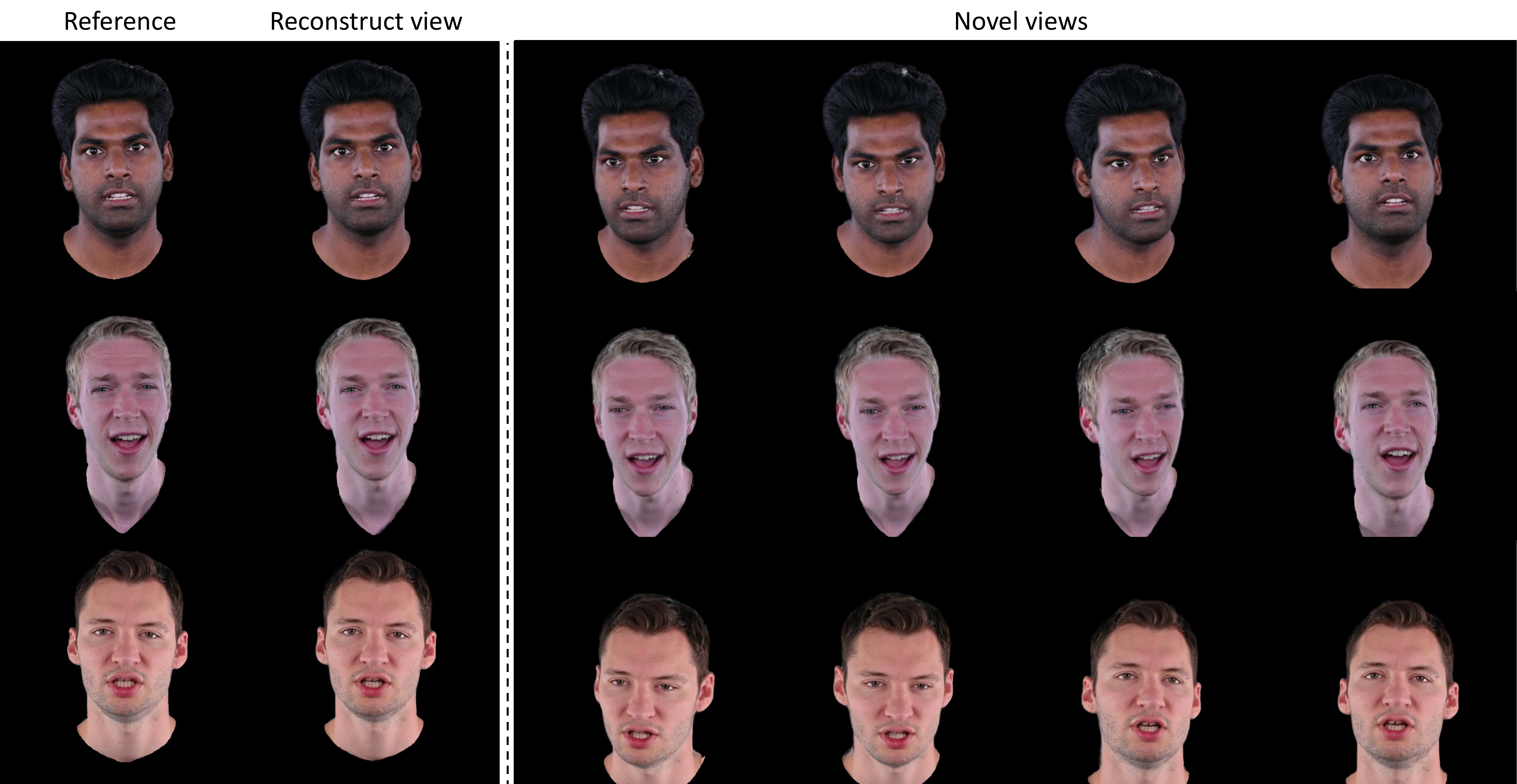}
  \caption{Visual results of novel view synthesis. In each row, the original view of the reconstructed subject is shown on the left, while the rendered novel views are on the right. Our method produces high-fidelity novel views with strong 3D consistency.
  }
  \label{show-novel-views}
\end{figure*}

\section{Applications}
\label{AdditionalApplications}
\subsection{Relighting}
We show the relighting results of the head illuminated by rotating environment maps in \cref{show-relight}. For each map, we extract the corresponding irradiance and prefiltered maps, applying them in the shading process (\cref{Appearance modeling}). HRAvatar achieves real-time rendering speed during relighting

For convenience during relighting, we use off-the-shelf tools to precompute the irradiance map and pre-filtered environment map from the environment map. Specifically, we use \href{https://github.com/dariomanesku/cmftStudio}{\textcolor{cyan}{CmftStudio}}, a tool commonly used in real-time rendering pipelines to process HDR images for image-based lighting. With CmftStudio, we extract the original environment map with a resolution of $1024\times512$ into an irradiance map of $512\times256$ and a pre-filtered environment map with 7 mipmaps, ranging from $1024\times512$ to $16\times8$.
\subsection{Material Editing}
By modeling the avatar's material properties for physical shading, we can easily edit the avatar’s materials. In \cref{show-material-editing}, we show material editing under new lighting conditions by gradually increasing the base Fresnel reflectance, which enhances the metallic effect and reduces diffuse reflection. As shown, higher reflectance results in stronger specular reflections, validating the effectiveness of our physically-based shading model.
\subsection{Novel Views Synthesis}
Although the 3D avatar is reconstructed from a monocular video, it can still render novel views. \cref{show-novel-views} shows the visual results of our method. As shown, HRAvatar renders novel views of the head with high 3D consistency and quality, preserving fine texture details.

\section{More Discussion}
\label{MoreDiscussion}
\subsection{Method Comparison}
\textbf{FLARE.} Similar to most relighting methods, both FLARE and our approach use a BRDF reflection model to account for environmental lighting on head appearance. The key distinction lies in the 3D representation: FLARE adopts a mesh-based approach, while we leverage 3D Gaussian Splatting (3DGS) and extend it with physically-based shading. We further overcome 3DGS's limitations in modeling normals and decoupling highlights from albedo. Moreover, our improved deformation model further enables higher-fidelity avatar reconstruction while achieving faster rendering compared to FLARE.

\noindent \textbf{3DGS-based.} \textit{GBS.} While both GBS and our method employ blendshapes to model positional displacements, we introduce: 1) learnable blend skinning for per-point rotations; 2) end-to-end training of an expression encoder to enhance tracking; and 3) a novel appearance model for better material decomposition and relighting. \textit{Other 3DGS-based.} Compared to other existing 3DGS-based monocular reconstruction methods, HRAvatar introduces a more flexible deformation method and employs an end-to-end trained expression encoder for more accurate expression capture, leading to superior reconstruction quality. Furthermore, we pioneer realistic, relightable monocular Gaussian head reconstruction. The main differences are summarized in \cref{key-aspects-tab}.

\subsection{Future improvements.} 
The extra computation from blendshapes, linear skinning, and shading slows down 3DGS rendering, but offloading these tasks to the GPU via CUDA could mitigate this. These challenges present opportunities for future improvement.

While albedo supervision from existing models reduces coupling to some extent, highlights may still be misattributed to properties like roughness or reflectance. Ideally, the same region, such as hair or skin, should have consistent material attributes. Introducing semantic information to guide and constrain material learning is a promising future direction.

\subsection{Ethical Considerations.} Creating realistic, controllable head avatars raises concerns about potential violations of portrait rights and privacy. It may also lead to identity theft and misuse in fraud. We strongly condemn any unauthorized use of this technology for illegal purposes. It's crucial to consider ethical implications in all applications of our method to prevent harm to the public.